\documentclass[sigconf,nonacm]{acmart}
\AtBeginDocument{%
  }

\makeatletter
\renewcommand\footnotetextcopyrightpermission[1]{}%
\makeatother

\usepackage{amsmath,amssymb,amsfonts}
\usepackage{xspace}
\usepackage{algorithmic}
\usepackage{algorithm}
\usepackage{graphicx}
\usepackage{textcomp}
\usepackage{xcolor}
\usepackage{booktabs}
\usepackage{multirow}
\usepackage{pifont}
\usepackage[inline]{enumitem}
\usepackage{balance}
\usepackage{subcaption}
\usepackage{hyperref}
\usepackage{cleveref}
\usepackage{colortbl}
\usepackage{makecell}
\usepackage{tikz}
\IfFileExists{twemojis.sty}{\usepackage{twemojis}}{\newcommand{\twemoji}[2][]{}}

\definecolor{darkgreen}{rgb}{0.0, 0.5, 0.0}
\definecolor{darkred}{rgb}{0.6, 0.0, 0.0}
\definecolor{darkorange}{rgb}{0.8, 0.4, 0.0}
\definecolor{lightblue}{rgb}{0.85, 0.92, 1.0}
\definecolor{oursrow}{rgb}{0.9, 0.95, 0.9}
\definecolor{myorange}{RGB}{255, 140, 0}
\definecolor{mygold}{RGB}{200, 150, 0}
\definecolor{myteal}{RGB}{0, 128, 128}
\definecolor{mycyan}{RGB}{0, 160, 200}
\definecolor{mypurple}{RGB}{128, 0, 128}
\definecolor{myviolet}{RGB}{138, 43, 226}

\definecolor{blockA}{HTML}{EBF5FB}      % Light blue-gray: Qwen1.5 rows
\definecolor{blockB}{HTML}{FFFFFF}      % White: DeepSeek rows
\definecolor{blockC}{HTML}{F0FFF0}      % Light green tint: GLM rows
\definecolor{blockD}{HTML}{FFF8F0}      % Light peach tint: Qwen3.5 rows
\definecolor{bestcell}{HTML}{D5F5E3}    % Mint green: best highlight
\definecolor{secondcell}{HTML}{FEF9E7}  % Pale yellow: second-best highlight
\definecolor{headtext}{HTML}{2C3E50}    % Dark blue-gray: header
\definecolor{methodsep}{HTML}{AEB6BF}   % Medium gray: group separator

\usepackage{xcolor}
\hypersetup{
  colorlinks,
  linkcolor={blue!70!green},
  citecolor={green!70!blue},
  urlcolor={orange!70!red}
}
\newcommand{\markyes}{\textcolor{darkgreen}{\ding{51}}}        % Checkmark: fully applies
\newcommand{\markno}{\textcolor{darkred}{\ding{55}}}           % Cross: does not apply
\newcommand{\markpartial}{\textcolor{mypurple}{\ding{119}}}  % Bullet: partially applies

\newcommand{\vect}[1]{\mathbf{#1}}
\newcommand{\mat}[1]{\mathbf{#1}}
\newcommand{\set}[1]{\mathcal{#1}}

\newcommand{\frob}[1]{\left\|#1\right\|_F}

\newcommand{\OurModel}{SEAL\xspace}
\newcommand{\OurMethodA}{SEAL\xspace}
\newcommand{\OurMethodB}{SEAL++\xspace}

\newcommand{\SE}{SE\xspace}
\newcommand{\RE}{RE\xspace}

\begin{document}

% --- arXiv preprint banner: venue notice on page 1 ---
\fancypagestyle{firstpagestyle}{%
  \fancyhf{}%
  \renewcommand{\headrulewidth}{0pt}%
  \renewcommand{\footrulewidth}{0pt}%
  \fancyhead[C]{\raisebox{0pt}[0pt][0pt]{\LARGE To Appear in ACM CCS 2026, November 2026}}%
  \fancyfoot[C]{\footnotesize\thepage}%
}

\title{
\texorpdfstring{\twemoji[height=0.8em]{seal}}{}
\OurModel: Reinforcing Global Safety in Mixture-of-Experts through \texorpdfstring{\underline{S}}{S}hared \texorpdfstring{\underline{E}}{E}xpert \texorpdfstring{\underline{A}}{A}\texorpdfstring{\underline{L}}{L}ignment}

\author{Qingyu Meng}
\email{q.meng@vu.nl}
\affiliation{%
  \institution{Vrije Universiteit Amsterdam}
  \city{Amsterdam}
  \country{Netherlands}}

\author{Yiwei Zha}
\affiliation{%
  \institution{Vrije Universiteit Amsterdam}
  \city{Amsterdam}
  \country{Netherlands}}

\author{Jiahuan Pei}
\affiliation{%
  \institution{Vrije Universiteit Amsterdam}
  \city{Amsterdam}
  \country{Netherlands}}

\author{Koen Hindriks}
\affiliation{%
  \institution{Vrije Universiteit Amsterdam}
  \city{Amsterdam}
  \country{Netherlands}}

\author{Herbert Bos}
\affiliation{%
  \institution{Vrije Universiteit Amsterdam}
  \city{Amsterdam}
  \country{Netherlands}}

\author{Min Chen}
\authornote{Corresponding author.}
\email{m.chen2@vu.nl}
\affiliation{%
  \institution{Vrije Universiteit Amsterdam}
  \city{Amsterdam}
  \country{Netherlands}}

\renewcommand{\shortauthors}{Meng et al.}

\begin{abstract}
Mixture-of-Experts (MoE) is a scaling architecture for large language models that activates only a small subset of expert modules per token, enabling massive parameter growth with nearly constant computation. Recent Hybrid MoE architecture adds \textit{shared experts} to capture consistently useful representations, further improving stability and generalization. MoE now powers many flagship open-source and commercial models, yet remains vulnerable to adversarial attacks. Specifically, sparse routing introduces a structural vulnerability: MoE safety hinges on which experts are activated, and adversaries can subvert this selection through jailbreak prompts, malicious fine-tuning, and weight-level pruning of safety-critical neurons. Existing defenses primarily focus on hardening the router, but an adversary may still manipulate or bypass the routing trajectory due to the routing process's nondeterministic nature, thereby collapsing the defense.

To cope with this problem, we first identify theoretically and empirically that shared expert, an always-activated component containing a small proportion of safety-critical neurons, can overcome the uncertainty of sparsely activated routing path and serve as a router-independent anchor to enhance global safety alignment. Based on this insight, we propose SEAL, a training-time parameter-efficient defense that produces a plug-and-play adapter attached to shared expert, and SEAL++, a variant that adds an orthogonal constraint preserving pre-existing safety subspaces during training. We evaluate SEAL and SEAL++ across six attack scenarios that combine three adversarial inputs (harmful prompting, jailbreak, malicious fine-tuning) with and without neuron pruning. SEAL reduces attack success rate (ASR) by up to 60\%, at a capability cost of at most 1.4\% on a five-benchmark average. Additionally, SEAL can seamlessly integrate with router-level defenses for stronger defense. Under this setting, we can further decrease the ASR to 4.8\%, more than five times lower than router-level defense alone. 
More broadly, our work demonstrates that shared experts constitute a unique defense surface overlooked by prior work, offering a complementary safety alignment pathway.
\end{abstract}




\maketitle

\begin{figure}[h]

  \centering
  \includegraphics[width=\columnwidth]{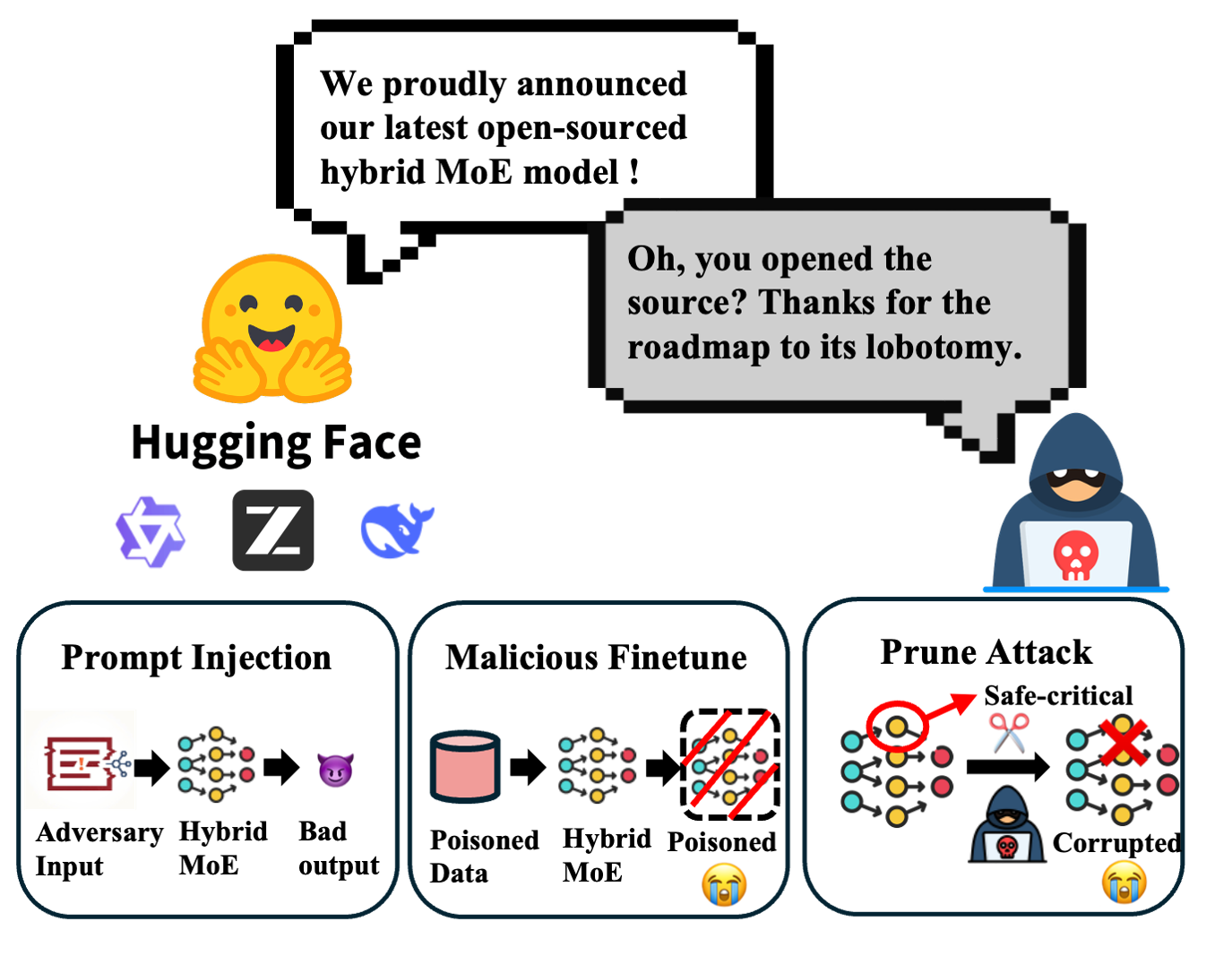}
  \caption{Three attack surfaces of hybrid MoE-based models.
  A malicious user can download any open-source model and apply three types of safety misalignment to bypass its safety guardrails.
  Prompt injection, malicious fine-tuning, and neuron pruning all exploit the same structural weakness: safety behavior depends on the experts selected.
  }
  \Description{Three vertical panels showing the three attack surfaces against a hybrid MoE model: prompt injection manipulating routing, malicious fine-tuning overwriting safety weights, and neuron-level pruning that disables safety-critical experts.}
  \label{fig:intro}
\end{figure}

 \section{Introduction}
\label{sec:intro}
Mixture-of-Experts (MoE) is a scaling strategy for large language models. It replaces selected feed-forward layers with parallel expert modules and a lightweight router that activates only a small subset of experts per input token. This design allows parameter counts to grow from billions to trillions while keeping per-token compute roughly constant. Currently, MoE architecture is the basis of many widely deployed language models, including GPT-OSS~\cite{openai2025gptoss120bgptoss20bmodel}, Llama~4~\cite{llama4-2025}, Qwen3.5~\cite{qwen2026-q35}, Nemotron-3~\cite{nemotron3-2025}, and Mixtral~\cite{jiang2024mixtralexperts}. 

Classical MoE activates only the top-$k$ of $N$ experts per token~\cite{fedus-switchtransformer}, so every expert competes for specialization and there is no principled place for knowledge that every token needs. Recent \emph{hybrid} MoE designs close this gap with \emph{shared experts}, modules that execute on every token regardless of routing~\cite{dai-etal-2024-deepseekmoe}, recognizing that some knowledge is too universal to leave to a router that cannot guarantee delivery. Since its introduction, the shared-expert design has propagated far beyond its original model demonstration and now appears in many frontier open-weight hybrid-MoE releases. Verified recent deployments include the DeepSeek V2/V3/R1/V4 family~\cite{deepseekai2026deepseekv4, deepseek-r1-2025}, the shared-expert Qwen MoE releases (Qwen1.5-MoE, Qwen2-MoE, Qwen3-Next, and
Qwen3.5)~\cite{yang2025qwen3technicalreport}, the GLM-4.5/4.6/4.7-Flash series~\cite{glm4.5-2025}, Kimi-K2~\cite{kimi2025k2}, Hunyuan-Large and Hunyuan-TurboS~\cite{sun2024hunyuanlargeopensourcemoemodel,
tencent2025hunyuanturbos}, Meta's Llama~4~\cite{llama4-2025}, and Google's Gemma~4
26B-A4B~\cite{google2026gemma4}. Meta, Google, and Nvidia each adopted this pattern for their most recent frontier MoE releases, consolidating the shared-expert design as the prevailing industrial choice for hybrid MoE and placing every modern open-weight deployment of this family within scope of a shared-expert-targeted defense.

\textbf{Attacks on MoE Safety.} 
Sparse routing exposes three distinct attack surfaces. 
An attacker can construct adversarial inputs ($\mathcal{A}_{\text{input}}$) to elicit harmful outputs by exploiting model behavior and manipulating routing distributions to bypass safety-critical experts~\cite{wei2023jailbroken, lai2025safex}. 
An attacker with direct access to model weights ($\mathcal{A}_{\text{weight}}$) can identify and prune safety-critical experts or neurons, disabling safety at inference time~\cite{wu2026gatebreaker, wu2026neurostrike}. 
An attacker can launch malicious finetuning attacks ($\mathcal{A}_{\text{train}}$) to misalign the model~\cite{qi2024finetuning}. 
All of these attacks, described in \autoref{fig:intro}, exploit the same structural vulnerability of MoEs: safety depends on which experts are selected, and an attacker who controls that selection can degrade safety.

\textbf{Defenses for MoEs.} Two families of MoE-specific defense have emerged, but neither fully addresses the existing vulnerability. \emph{Router-level defenses}~\cite{kim2026defendingmoellmsharmful, fayyaz2026steeringmoellmsexpert} harden the gating decision through KL-regularized routing or inference-time logit steering, and so depend on the routing mechanism to be reliable. \emph{Routed-expert interventions}~\cite{liang2026rasaroutingawaresafetyalignment} fine-tune the safety-critical routed experts themselves, but can only raise safety on tokens the router still delivers to those experts. In either case the router is the central guarded part of the defense, so neither family hardens against adversaries who can manipulate routing or ignore it.

Beyond the defense strategies discussed above, the inherent always-active nature of shared experts opens up a new defense surface in MoE architectures. Unlike routed experts, whose activation is controlled by the router and therefore subject to manipulation, shared experts remain consistently engaged regardless of how an attacker compromises the routing mechanism. 
Despite this natural advantage, no prior work has studied or explored the shared expert as a dedicated safety component, leaving a significant gap that our work is the first to address.

We systematically investigate the shared-expert parameter subspace as a safety surface across four hybrid MoE architectures, and find that shared experts carry safety-relevant neurons at per-parameter density comparable to routed experts (shared-to-routed density ratio between 0.63 and 0.89 across the four models), yet, unlike routed experts, they execute unconditionally on every token, so their safety contribution is active on every forward pass regardless of routing state. Despite this combination of meaningful density and guaranteed execution, no existing defense targets the shared-expert subspace, leaving an architecturally guaranteed safety surface unexploited. On the basis of this finding, we propose \OurModel (\textbf{S}hared \textbf{E}xpert \textbf{AL}ignment), a training-time framework that reinforces shared experts for router-independent safety enforcement.
\OurMethodA trains Direct Preference Optimization (DPO)-based Low-Rank Adaptation (LoRA) adapters restricted to shared-expert projections, modifying only ${\sim}0.25\%$ of model parameters.
\OurMethodB extends \OurMethodA with an orthogonal constraint on the profiled
safety subspace, which prevents DPO training from overwriting the pre-existing
safety-critical directions that make safety neurons robust to pruning.
\OurModel operates exclusively on shared-expert parameters, a subset distinct from the routing parameters targeted by existing router-level defenses~\cite{liang2026rasaroutingawaresafetyalignment, kim2026defendingmoellmsharmful}, so the two families do not interfere and can be directly composed. Deploying \OurModel on top of routing-level protections therefore yields defense in depth, where router-level methods guard the routing mechanism and \OurModel anchors safety in a parameter region the router cannot reach.

Our contributions are as follows:
\begin{itemize}[leftmargin=*,nosep]
    \item We identify \emph{shared experts} as a previously unexplored, router-independent safety surface in hybrid MoE architectures. Empirical analysis across four open-weight models shows that shared-expert parameters carry safety-relevant neurons at a per-parameter density comparable to routed experts while being activated on every token, establishing the concept of \emph{safety shared experts} as dedicated defense anchors.
    \item We propose \OurMethodA, a training-time defense that applies DPO-based LoRA adapters exclusively to shared expert projections, modifying only 0.06--0.25\% of total model parameters and encoding router-independent safety representations.
    \item We extend \OurMethodA with \OurMethodB, which adds an orthogonal constraint that prevents DPO training from overwriting pre-existing safety-critical subspaces, further raising the cost of post-deployment neuron-level pruning attacks.
    \item We conduct a comprehensive evaluation of \OurMethodA and \OurMethodB against the three threat-model scenarios and compound attacks that combine prompt injection, malicious fine-tuning, and neuron-level pruning, and additionally show that \OurMethodB composes with router-level defenses to further strengthen pruning robustness at negligible capability cost.
\end{itemize}

\begin{figure*}[ht]
  \centering
  \includegraphics[width=0.95\textwidth]{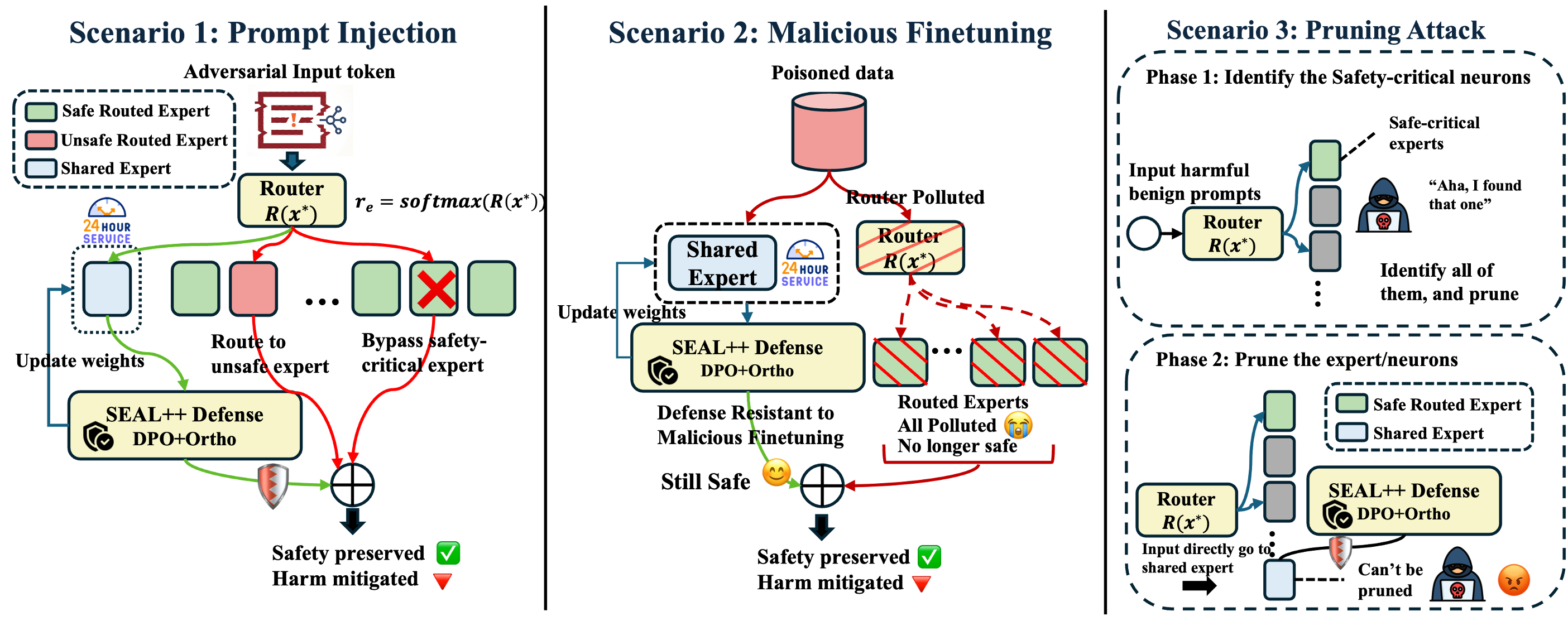}
  \caption{Threat model. Three adversaries with escalating privileges ($\mathcal{A}_{\text{input}}$, $\mathcal{A}_{\text{train}}$, $\mathcal{A}_{\text{weight}}$) target the MoE safety surface through prompt injection, malicious fine-tuning, and parameter-level pruning, respectively. \OurModel restricts gradient updates to shared-expert parameters, producing a router-independent defense surface that persists across all three scenarios.}
  \Description{Diagram of the threat model with three adversary types (input, train, weight) attacking a hybrid MoE architecture, showing how each scenario compromises a different parameter region while shared experts remain the router-independent defense surface for \OurModel.}
  \label{fig:threat}
\end{figure*}

\section{Preliminaries}
\label{sec:prelim}

\subsection{Transformer and MoE Architectures}
\label{subsec:llm_bg}

A large language model (LLM) parameterized by $\theta$ generates tokens autoregressively through a stack of $L$ Transformer layers. At layer $l$, a self-attention sublayer captures inter-token dependencies, and a feed-forward network (FFN) sublayer transforms the resulting hidden state $\vect{h}^{(l)} \in \mathbb{R}^d$. In \emph{dense} Transformers, every FFN parameter participates in every forward pass, coupling the total parameter count directly to per-token compute.

Mixture-of-Experts (MoE) architectures decouple the parameters from the computation by replacing selected FFN sublayers with $E$ parallel \emph{expert} modules and a lightweight \emph{router} (gating network) $R\!: \mathbb{R}^d \!\to\! \{0,1\}^E$. Given an input token representation $\vect{x} \in \mathbb{R}^d$, the router assigns scores to all experts and activates only the top-$k$ ($k \ll E$, typically 2--8). The remaining experts stay dormant. This conditional activation allows parameter counts to scale from billions to trillions while keeping per-token FLOPs roughly constant. Crucially, different tokens traverse different subsets of parameters, making model behavior dependent on routing dynamics.

\subsection{Hybrid MoE and Shared Experts}
\label{subsec:moe_bg}
A \emph{Hybrid MoE} layer at depth $l$ partitions its FFN capacity into $S$ \emph{shared experts} (\SE) $\{g_s\}_{s=1}^{S}$ and $E$ \emph{routed experts} (\RE) $\{f_e\}_{e=1}^{E}$.
Each expert implements a gated feed-forward network with three projection matrices (\texttt{gate\_proj}, \texttt{up\_proj}, \texttt{down\_\allowbreak{}proj}).
For input hidden state $\vect{x} \in \mathbb{R}^d$, the layer output is:
\begin{equation}
h(\vect{x}) = \underbrace{\sum_{s=1}^{S} g_s(\vect{x})}_{\text{shared: always active}} + \underbrace{\sum_{e=1}^{E} r_e(\vect{x}) \cdot f_e(\vect{x})}_{\text{routed: conditionally active}}
\label{eq:moe_layer}
\end{equation}
where $r_e(\vect{x}) \in \{0,1\}$ denotes the binary top-$k$ routing decision, satisfying $\sum_{e=1}^{E} r_e(\vect{x}) = k$ for a fixed $k \ll E$.

Shared experts were introduced by Dai et al.~\cite{dai-etal-2024-deepseekmoe} to aggregate common knowledge that would otherwise be repetitively encoded across multiple routed experts, thereby improving expert specialization and reducing inter-expert redundancy. Due to the fact that shared experts execute unconditionally on every token (\autoref{eq:moe_layer}), their contribution to the layer output persists regardless of the router's top-$k$ selection. Routed experts, by contrast, only activate when selected by the router. This structural asymmetry between unconditional and conditional expert execution creates a unique defense surface in hybrid MoE architectures. This routing independence is the foundation for the approach.
\autoref{tab:notation_appendix} provides a complete notation reference.

\section{Threat Model}
\label{sec:threat}

We formalize three adversarial scenarios in order of escalating attacker privileges, depicted in \autoref{fig:threat}. Let $\theta$ denote model parameters and $\mathcal{I}$ denote the input interface.

\begin{definition}[Adversary Capability]
\label{def:adversary}
\[
\left\{
\begin{tabular}
{@{}l@{\hspace{1em}}l@{\hspace{1em}}l@{}}
\xspace $\mathcal{A}_{\text{input}}$  & Construct adversarial inputs $x \in \mathcal{I}$ to query model. \\
\xspace $\mathcal{A}_{\text{train}}$  & Train model with attacker-controlled data. \\
\xspace $\mathcal{A}_{\text{weight}}$ & Read and edit model parameters $\theta$.
\end{tabular}
\right.
\]
\end{definition}

\paragraph{Scenario A (Prompt Injection Attacks).}
Adversary with $\mathcal{A}_{\text{input}}$ capability has black-box query access: the attacker can submit arbitrary inputs $x \in \mathcal{I}$ and observe outputs, with limited knowledge of model architecture, weights, or internal routing decisions~\cite{wei2023jailbroken, shen2024dan}. 
The attacker constructs adversarial prompts, including template-based jailbreaks and human-crafted multi-turn manipulation, to steer model behavior toward harmful outputs. These prompts exploit learned response patterns and may implicitly shift routing distributions away from safety-critical experts.
A successful jailbreak causes the model to generate compliant responses to harmful instructions that its safety alignment would normally refuse. This scenario represents the most common real-world attack surface. Because shared-expert augmentation operates entirely at the parameter level, it composes naturally with input-space defenses such as prompt classifiers and perplexity filters for layered protection.

\paragraph{Scenario B (Malicious Fine-Tuning).}
Adversary with $\mathcal{A}_{\text{train}}$ capability has access to model weights and a training framework, enabling supervised fine-tuning on attacker-controlled data~\cite{qi2024finetuning}. This models a downstream user or supply-chain actor who weaponizes open weights models. 
The attacker fine-tunes the full model on harmful instruction-response pairs via standard supervised learning. Unlike Scenario~A, the adversary can modify parameters through gradient updates, including routed expert weights which are not primarily targeted by shared-expert defense.
Malicious fine-tuning overrides safety alignment by corrupting learned refusal behavior across the parameter space. This scenario reveals the extent to which safety encoding concentrated in shared experts persists under gradient-based attacks that modify the entire parameter space.

\paragraph{Scenario C (Pruning Attacks).}
Adversary with $\mathcal{A}_{\text{weight}}$ capability obtains full model weights and can read and edit parameters directly, but cannot invoke gradient-based training~\cite{wu2026gatebreaker, wu2026neurostrike}. The key distinction from $\mathcal{A}_{\text{train}}$ is operational, as weight-level editing requires no training framework or curated fine-tuning data, and its GPU cost is limited to forward-pass activation profiling rather than full gradient-based training, making it feasible in inference-only deployments. 
The attacker profiles activations on harmful and benign datasets, identifies safety-critical neurons, and masks them at inference time. This covers a malicious operator performing targeted weight surgery on a locally hosted model, a red-team auditor probing safety boundaries via selective neuron ablation, or a compression pipeline that removes safety-critical neurons during post-training sparsification. Recent work has shown that modifying only a small fraction of neurons can substantially degrade safety behavior in MoE models~\cite{wu2026gatebreaker, wu2026neurostrike}, making this scenario a practical and low-cost surface.
Selective neuron pruning disables safety behavior at inference time while preserving general capability. This is a particularly low-cost attack, and forms the primary evaluation axis of this work. A shared-expert defense raises the pruning cost by concentrating and reinforcing safety representations in the always-active parameter subspace.

\paragraph{Design Oracle.}
Given the three scenarios depicted above, a satisfactory defense for hybrid MoE safety alignment should meet four requirements.
\emph{(D1) Non-privileged runtime access.} 
The defender operates at training time on a benign development infrastructure, with white-box access to the model, a curated preference dataset of safe-versus-unsafe responses, and the ability to attach lightweight parameter-efficient modules. 
\emph{(D2) Utility preservation.} The defense must not measurably degrade general capability, with capability drift on standard benchmarks bounded to within a small tolerance (we target $\leq\!5\%$).
\emph{(D3) Robustness.} The defense must remain effective under $\mathcal{A}_{\text{input}}$, $\mathcal{A}_{\text{train}}$, and $\mathcal{A}_{\text{weight}}$.
\emph{(D4) Affordable cost with minimum pipeline modification.} The defense must fit into standard MoE fine-tuning pipelines without altering the router, routed experts, or serving infrastructure.
\OurModel satisfies (D1) by restricting parameter updates to shared-expert parameters using DPO-based LoRA adapters, (D2) by showing about only 1\% perturbation to model capability, (D3) by anchoring safety in an always-active subspace that persists under prompt injection, malicious fine-tuning, and pruning attacks, and (D4) by leaving router and routed-expert weights untouched. \OurModel thus provides \emph{resilience} rather than integrity, complementing weight checksums: a checksum detects whether weights have changed, while shared-expert augmentation ensures the model remains safer even after modification, providing defense in depth when integrity checks are absent or bypassed. \autoref{alg:seal} and \autoref{alg:sealplus} formalise the two training variants.

\section{Our Method}
\label{sec:method}

In this section, we first motivate the use of shared experts as a safety augmentation surface in \autoref{subsec:motivation}.
Then we perform a feasibility analysis in \autoref{subsec:param_cost}.
We propose shared-expert-based defense method in \autoref{subsec:seal} and its variant in \autoref{subsec:sealplus}.

\subsection{Motivation}
\label{subsec:motivation}

Four complementary arguments support shared experts as a suitable target for safety augmentation in hybrid MoE architectures.

\paragraph{Property 1: Unconditional Execution Guarantee.} 
By \autoref{eq:moe_layer}, the shared expert term $\sum_{s=1}^{S} g_s(\vect{x})$ executes for every input $\vect{x}$ regardless of the 
routing function $R$. 
Safety behavior encoded in shared expert parameters $\Theta_{\text{shared}} = \{\theta_{g_s}\}_{s=1}^{S}$ 
therefore activates on every token, even when the router is fully compromised. 
This guarantee is complementary to router-dependent defenses~\cite{liang2026rasaroutingawaresafetyalignment, kim2026defendingmoellmsharmful} that strengthen safety \emph{when routing is intact} 
: shared-expert augmentation provides an additional fallback that holds \emph{regardless of routing state}, and combining both could yield a globally stronger defense than either alone.

\begin{figure}[t]
  \centering
  \includegraphics[width=\columnwidth]{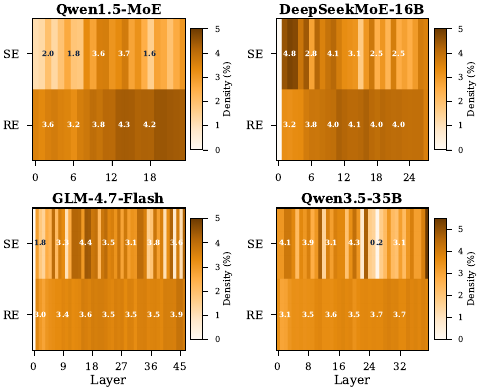}
  \caption{Per-layer safety neuron density (\%) distribution. 
  Shared experts (SE) hold a relatively smaller absolute proportion, yet their unconditional execution makes each safety neuron active on every token, while routed-expert (RE) safety neurons contribute only when selected by the router.}
  \Description{Four heatmaps (Qwen1.5, DeepSeek, GLM, Qwen3.5) plotting per-layer safety neuron density for shared and routed experts on an orange colour scale, showing that shared experts hold a small but consistent fraction of safety neurons across every layer of every architecture.}
  \label{fig:safety_heatmap}
\end{figure}

\paragraph{Property 2: Attack Surface Reduction.}
The attacks discussed in \autoref{sec:threat} share a common pattern: they exploit the separation between safety-carrying parameters and the routing path. Pruning attacks~\cite{wu2026gatebreaker, wu2026neurostrike} show that disabling 1--5\% of neurons suffices to collapse safety alignment, with shared experts carrying disproportionate safety impact relative to their size. Routing-level attacks~\cite{lai2025safex} bypass safety-critical experts through distribution shifts. Placing safety augmentation in shared experts forces the attacker to target a component that is architecturally guaranteed to execute, mitigating routing bypass as an attack and raising the effort required for safety removal. 
\autoref{fig:safety_heatmap} quantifies this asymmetry across four
architectures. Safety neurons are present in both expert types, with a shared-to-routed per-module density ratio of $0.63$ (Qwen1.5), $0.88$
(DeepSeek), $0.89$ (GLM), and $0.86$ (Qwen3.5). Despite the slightly lower per-module density on shared experts, the guaranteed execution from Property~1 ensures every safety neuron in this subspace contributes to every
forward pass, making it a reliable anchor for safety augmentation.

\paragraph{Property 3: Defense Isolation.}
Shared experts constitute a small fraction of total model parameters (\autoref{tab:param_cost}). 
Restricting updates to shared experts avoids modifying routing weights, ensuring the defense does not introduce new attack surface through routing perturbation.
Routed experts, which encode the majority of task-specific knowledge~\cite{dai-etal-2024-deepseekmoe}, remain frozen during the parameter updating process (\autoref{tab:param_cost}). 
In MoE architectures, unconstrained full-model alignment risks representation collapse~\cite{chi2022representation_collapse} through routing destabilization, further motivating the restriction to shared experts.

\paragraph{Property 4: Representation Isolation.} 
In MoE training, loss gradients decompose into an unconditional component through shared experts and a routing-conditional component through routed experts~\cite{dai-etal-2024-deepseekmoe, lo-etal-2025-closer}. This implicit decoupling means safety features injected into shared experts occupy a subspace partially isolated from task-specific knowledge in routed experts, reducing gradient conflict during alignment training~\cite{wang2024esft, liu2024perft, dou2024loramoe}.

\subsection{Parameter Cost}
\label{subsec:param_cost}
We instantiate \OurModel on four representative hybrid-MoE models that used throughout the paper, chosen to cover the core axes of the hybrid-MoE design space. 
For each of the chosen model, we restrict LoRA adaptation~\cite{hu2022lora} to the three projection matrices (\texttt{gate\_proj}, \texttt{up\_proj}, \texttt{down\_proj}) in every shared expert across all MoE layers. 
For a projection matrix $\mat{W} \in \mathbb{R}^{d_{\text{out}} \times d_{\text{in}}}$ with LoRA rank $r$, the adapter introduces $r \times (d_{\text{in}} + d_{\text{out}})$ trainable parameters (matrices $\mat{A} \in \mathbb{R}^{r \times d_{\text{in}}}$ and $\mat{B} \in \mathbb{R}^{d_{\text{out}} \times r}$). The total trainable parameter count is:

\begin{equation}
|\Theta_{\text{train}}| = \underbrace{L_{\text{MoE}}}_{\text{MoE layers}} \times \underbrace{S}_{\text{shared experts}} \times \underbrace{3}_{\text{projections}} \times \underbrace{r \cdot (d + d_{\text{FFN}})}_{\text{per adapter}}
\label{eq:param_cost}
\end{equation}

\noindent where $d$ is the hidden dimension, $d_{\text{FFN}}$ the intermediate (FFN) dimension of each shared expert, $L_{\text{MoE}}$ the number of MoE layers, and $S$ the number of shared experts per layer. All three projections share the same LoRA rank $r = 64$ and scaling factor $\alpha = 128$.

\begin{table}[t]
\centering
\caption{Parameter cost of shared expert training. 
Total: total parameter count; 
$L_{\text{MoE}}$: number of MoE layers; 
$S$ : shared experts per layer; 
$d_{\text{FFN}}$: shared expert FFN dimension; 
Ratio: trainable fraction of total parameters.
}
\label{tab:param_cost}
\small
\setlength{\tabcolsep}{2.5pt}
\begin{tabular}{@{}l r r r r r r@{}}
\toprule
\textbf{Model} & \textbf{Total} & \textbf{$L_{\text{MoE}}$} & \textbf{$S$ } & \boldmath$d_{\text{FFN}}$ & \textbf{Trainable} & \textbf{Ratio} \\
\midrule
Qwen1.5-MoE-A2.7B\footnotemark[1]  & 14.3B & 24 & 1 & 5{,}632 & 35.4M & 0.25\% \\
DeepSeekMoE-16B     & 16.4B & 27 & 2 & 1{,}408 & 35.8M & 0.22\% \\
GLM-4.7-Flash       &   30B & 46 & 1 & 1{,}536 & 31.7M & 0.11\% \\
Qwen3.5-35B-A3B\footnotemark[2]     &   35B & 40 & 1 &     512 & 19.7M & 0.06\% \\
\bottomrule
\end{tabular}
\end{table}
\footnotetext[1]{Conceptually 4 shared experts bundled as one wide FFN ($d_{\text{FFN}}{=}4{\times}1{,}408$).}

\autoref{tab:param_cost} reports the concrete costs. 
Across all models, the trainable budget remains below 36M parameters (at most 0.25\% of total model size), with over 99.75\% of parameters frozen throughout training.
The variation in trainable parameter count reflects differences in shared expert FFN dimensions and counts across models.

\begin{figure*}
  \centering
  \includegraphics[width=\textwidth]{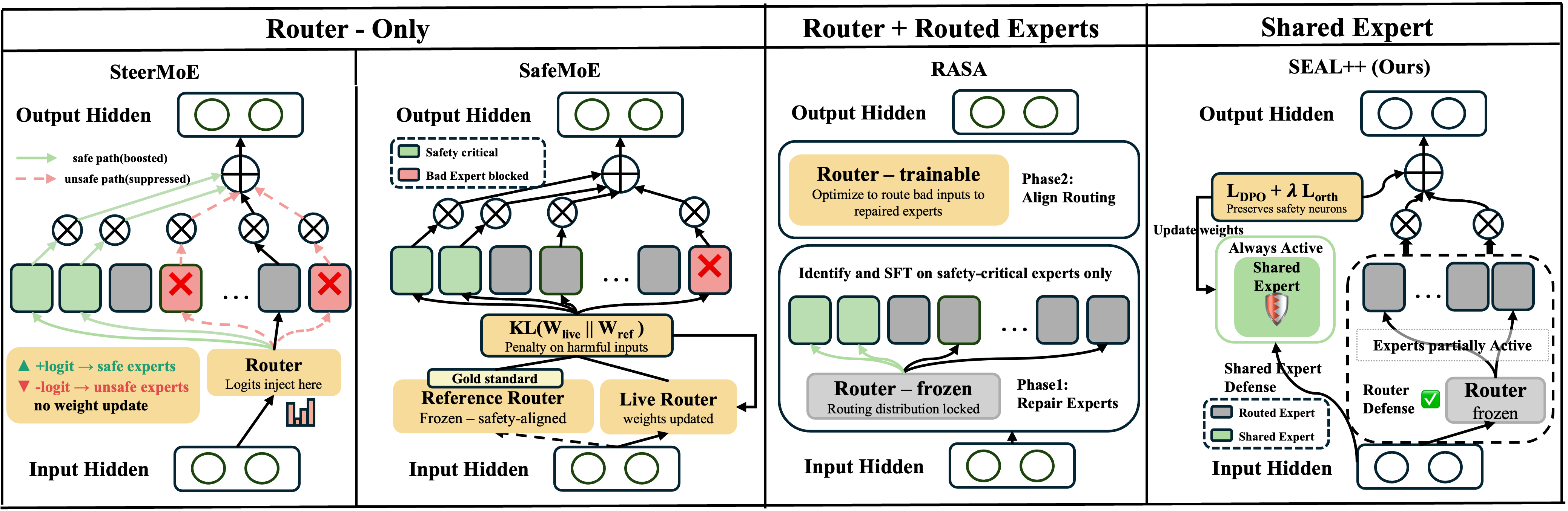}
  \caption{Defense surfaces in MoE safety alignment. \emph{Router-only}: SteerMoE~\cite{fayyaz2026steeringmoellmsexpert} steers expert logits at inference; SafeMoE~\cite{kim2026defendingmoellmsharmful} regularizes the live router against a frozen safety-aligned reference. \emph{Router plus routed-expert}: RASA~\cite{liang2026rasaroutingawaresafetyalignment} repairs safety-critical routed experts and re-aligns the router toward them. \OurMethodB (ours) leaves both the router and routed experts untouched, training only the always-active shared expert with DPO plus an orthogonal constraint $\mathcal{L}_{\text{orth}}$, placing the defense on a parameter subset whose execution is router-independent.}
  \Description{Three-column schematic comparing defense loci in MoE safety: Router-only defenses (SteerMoE, SafeMoE) operate on the router; Router-plus-routed-expert (RASA) edits both; our SEAL / SEAL++ variant trains only the always-active shared expert and leaves the router and routed experts untouched.}
  \label{fig:defense_surface}
\end{figure*}

The preceding analysis establishes that shared experts possess a unique combination of security properties: unconditional execution that survives routing compromise, a compact parameter count (0.06--0.25\%) that lowers the defense's cost, and representational isolation from task-specific knowledge in routed experts. \emph{Targeting shared experts for safety augmentation is therefore an architectural choice}, one that converts a structural invariant of hybrid MoE design into a concrete defense mechanism. We now describe how \OurModel exploits this surface. The framework comprises two variants: \OurMethodA (\autoref{subsec:seal}), which trains DPO-based LoRA adapters restricted to shared experts, and \OurMethodB (\autoref{subsec:sealplus}), which adds an orthogonal constraint that preserves existing safety-critical neuron subspaces during training. Both operate at training time and incur zero inference overhead.
We begin with safety-critical neuron identification (\autoref{subsec:profiling}), then
present each method in turn.

\subsection{Safety-critical Neurons Identification}
\label{subsec:profiling}

We identify safety-critical neurons in shared experts and construct a protection mask that defines the constrained dimensions for \OurMethodB. 
The procedure repurposes the activation differential analysis from neuron pruning attacks~\cite{wu2026gatebreaker} for defense: the neurons identified by the attacker to \emph{locate} safety-critical parameters are the same neuron sets we \emph{protect}.

For shared expert module $(l,s)$, we collect activations on harmful prompt dataset $\set{D}_h$ and benign prompt dataset $\set{D}_b$ at the intermediate representation. 
The per-neuron activation differential $\Delta a_j^{(l,s)}$ (\autoref{eq:act_diff} in \autoref{app:encoding}) quantifies each neuron's preferential activation on harmful content. 
Safety-critical neurons are identified via $z$-score thresholding:
\begin{equation}
\set{N}_{\text{safe}}^{(l,s)} = \left\{ j : \frac{|\Delta a_j^{(l,s)}| - \mu_{\Delta}}{\sigma_{\Delta}} > \zeta \right\}, \quad \zeta = 2.0
\label{eq:safety_id}
\end{equation}
The safety module $\mat{C}_{\text{SE}}^{(l,s)}$ is a diagonal matrix that masks the identified safety neurons:
\begin{equation}
[\mat{C}_{\text{SE}}^{(l,s)}]_{jj} = \begin{cases} 0 & \text{if } j \in \set{N}_{\text{safe}}^{(l,s)} \\ 1 & \text{otherwise} \end{cases}
\label{eq:cse}
\end{equation}

By construction, $\mat{C}_{\text{SE}}$ is a binary diagonal mask satisfying $\mat{C}_{\text{SE}}^2 = \mat{C}_{\text{SE}}$ (idempotent). 
The complementary matrix $\mat{I} - \mat{C}_{\text{SE}}$ selects precisely the safety-critical neuron dimensions. 
Since safety neurons typically constitute 1\% to 3\% of the intermediate representation~\cite{chen2025towards, wu2026gatebreaker}, the constraint removes a small fraction of the update space while protecting the most safety-relevant dimensions.
The profiling requires approximately 300 harmful and 300 benign prompts~\cite{li2025salora} and runs once per model architecture. The resulting matrices are precomputed and frozen during all subsequent training.

\subsection{\texorpdfstring{\OurMethodA: \underline{S}hared \underline{E}xpert \underline{A}\underline{L}ignment}{\OurMethodA: Shared Expert ALignment}}
\label{subsec:seal}

\begin{algorithm}[t]
\caption{\OurMethodA Training Procedure}
\label{alg:seal}
\small
\begin{algorithmic}[1]
\REQUIRE MoE architecture $\mathcal{M}$ with shared expert module set $\set{M}$
\REQUIRE DPO training data $\set{D} = \{(x_i, y_{w,i}, y_{l,i})\}_{i=1}^{N}$
\ENSURE Trained adapter $\theta^*$
\STATE Attach LoRA adapters $\mat{A}^{(l,s)} \!\in\! \mathbb{R}^{r \times d}$, $\mat{B}^{(l,s)} \!\in\! \mathbb{R}^{d_{\text{out}} \times r}$ to each projection (\texttt{gate\_proj}, \texttt{up\_proj}, \texttt{down\_proj}) in every $(l,s) \in \set{M}$
\STATE Freeze all parameters except $\theta = \{(\mat{A}^{(l,s)}, \mat{B}^{(l,s)})\}_{(l,s) \in \set{M}}$
\STATE $\pi_{\text{ref}} \leftarrow \mathcal{M}|_{\mat{BA}=\mat{0}}$ \COMMENT{implicit: disable adapters}
\FOR{each mini-batch $\mathcal{B} \subset \set{D}$}
\STATE Compute $\mathcal{L}_{\text{DPO}}$ on $\mathcal{B}$ using $\pi_\theta$ and $\pi_{\text{ref}}$ \COMMENT{\autoref{eq:dpo}}
\STATE $\theta \leftarrow \theta - \eta \, \nabla_\theta \mathcal{L}_{\text{DPO}}$
\ENDFOR
\RETURN $\theta^* \leftarrow \theta$
\end{algorithmic}
\end{algorithm}

\OurMethodA applies low-rank adaptation~\cite{hu2022lora} exclusively to shared expert projection layers, while all other parameters remain frozen. For shared expert projection matrix $\mat{W}^{(l,s)} \in \mathbb{R}^{d_{\text{out}} \times d}$, we attach adapters $\mat{B} \in \mathbb{R}^{d_{\text{out}} \times r}$ and $\mat{A} \in \mathbb{R}^{r \times d}$ with $r \ll \min(d_{\text{out}}, d)$, producing effective weight:
\begin{equation}
\mat{W}' = \mat{W} + \mat{B}\mat{A}
\label{eq:selora}
\end{equation}

Training optimizes $\mat{A}$ and $\mat{B}$ via Direct Preference Optimization (DPO)~\cite{rafailov-dpo} on safety preference data, which learns from human-annotated $(x, y_w, y_l)$ triples of prompt, chosen (safe) response, and rejected (unsafe/harmful) response (\autoref{eq:dpo} in \autoref{subsec:sealplus}). The low-rank parameterization serves two purposes beyond parameter efficiency: (1)~it provides an implicit reference policy for DPO, since disabling the adapters ($\mat{B}\mat{A} = \mat{0}$) recovers the frozen base model $\pi_{\text{ref}}$, eliminating the need for a separate model copy; (2)~the explicit adapter matrices $\mat{B}$ and $\mat{A}$ expose the update $\Delta\mat{W}$ in closed form, enabling the orthogonal constraint in \OurMethodB (\autoref{eq:orth_loss}). No orthogonality or safety-subspace constraint is imposed in \OurMethodA. It provides a natural reference point for evaluating the additional effect of the orthogonal constraint in \OurMethodB.

While \OurMethodA demonstrates that shared-expert DPO alone provides substantial defense, unconstrained gradient updates may unexpectedly impair the pre-existing safety representations that make safety neurons identifiable and robust to pruning. \OurMethodB seeks to address this gap by adding a geometric constraint that preserves these representations during training. \autoref{alg:seal} summarizes the \OurMethodA training procedure; the \OurMethodB extension in \autoref{alg:sealplus} adds the orthogonal-constraint step.

\subsection{\texorpdfstring{\OurMethodB: \underline{S}hared \underline{E}xpert \underline{A}\underline{L}ignment with Orthogonal Constraint}{\OurMethodB: Shared Expert Aligned LoRA with Orthogonal Constraint}}
\label{subsec:sealplus}

A growing body of work has shown that fine-tuning, even when intended to reinforce safety, can inadvertently disrupt a language model's pre-existing safety alignment, overwriting the refusal-relevant subspace of the base model and leaving the resulting model behaviorally compliant but representationally fragile under downstream attacks~\cite{qi2024finetuning, li2025salora, SafeLoRA}. \OurMethodB is designed to counter this potential side effect in the hybrid MoE setting.

\OurMethodB extends \OurMethodA with an orthogonal constraint that penalizes LoRA updates projecting onto the identified safety subspaces. The constraint prevents DPO training from overwriting the pre-existing safety representations that \OurMethodA builds upon, so gradient updates that improve safety behavior no longer erode the geometric structure that makes safety neurons identifiable and robust to pruning. Preserving these safety-critical representations during training raises the cost of post-deployment neuron-level pruning attacks (Scenario~C) and strengthens robustness to adversarial prompts (Scenario~A).

\subsubsection{Constrained Parameter Update}

For each shared expert module $(l,s) \in \set{M}$, the LoRA update $\Delta\mat{W} = \mat{B}^{(l,s)}\mat{A}^{(l,s)}$ decomposes into a safety-aligned component and a safety-orthogonal component:
\begin{equation}
\mat{B}\mat{A} = \underbrace{(\mat{B}\mat{A})(\mat{I} - \mat{C}_{\text{SE}})}_{\text{safety-aligned (penalized)}} + \underbrace{(\mat{B}\mat{A})\,\mat{C}_{\text{SE}}}_{\text{safety-orthogonal (allowed)}}
\label{eq:decomp}
\end{equation}
Right-multiplication by $\mat{I} - \mat{C}_{\text{SE}}$ (\autoref{eq:cse}) retains only the columns of $\mat{B}\mat{A}$ that correspond to safety neuron indices, projecting the update onto the safety-critical coordinate subspace. The orthogonal constraint loss penalizes the safety-aligned component:
\begin{equation}
\mathcal{L}_{\text{orth}} = \frac{1}{|\set{M}|} \sum_{(l,s) \in \set{M}} \frob{\mat{B}^{(l,s)} \mat{A}^{(l,s)} \cdot (\mat{I} - \mat{C}_{\text{SE}}^{(l,s)})}^2
\label{eq:orth_loss}
\end{equation}
Minimizing $\mathcal{L}_{\text{orth}}$ drives updates toward the $(d - r_s)$-dimensional safety-orthogonal subspace, where $r_s = |\set{N}_{\text{safe}}^{(l,s)}|$ denotes the number of safety neurons in layer $l$. 
The safety neurons typically constitute 1--3\% of the intermediate dimension, imposing minimal restriction on task learning while protecting safety-critical directions.

\subsubsection{Training Objective}

The complete \OurMethodB objective combines DPO~\cite{rafailov-dpo} with the orthogonal constraint:
\begin{equation}
\mathcal{L}_{\text{SEAL++}} = \mathcal{L}_{\text{DPO}} + \lambda_{\text{orth}} \cdot
\mathcal{L}_{\text{orth}}
\label{eq:total_loss}
\end{equation}
where $\lambda_{\text{orth}} > 0$ controls constraint strength (default $\lambda_{\text{orth}} = 0.1$). The DPO loss optimizes preference alignment on safety data:
\begin{equation}
\mathcal{L}_{\text{DPO}} = -\mathbb{E}_{(x,y_w,y_l)} \left[ \log \sigma \!\left( \beta \left( \log \frac{\pi_\theta(y_w|x)}{\pi_{\text{ref}}(y_w|x)} - \log \frac{\pi_\theta(y_l|x)}{\pi_{\text{ref}}(y_l|x)} \right) \right) \right]
\label{eq:dpo}
\end{equation}
Here $(x, y_w, y_l)$ is a preference triple consisting of prompt $x$, chosen response $y_w$, and rejected response $y_l$. $\pi_\theta$ denotes the policy model (trainable LoRA parameters $\theta$); $\pi_{\text{ref}}$ denotes the reference policy (frozen base model). The temperature $\beta > 0$ scales the implicit reward difference (default $\beta = 0.1$).\footnote{We adopt the standard DPO setting from~\cite{rafailov-dpo}, which balances reward sensitivity against training stability in low-rank fine-tuning.}

\paragraph{Why DPO on shared experts.}
DPO's margin objective between safe and unsafe responses, applied only to shared-expert parameters, could sharpen the geometric separation between harmful and benign activations in the unconditional path. Unlike full-model DPO approaches~\cite{secalign} that target input-level defense across all parameters, restricting updates to shared experts encodes the safety signal in router-independent parameters, so the learned preference persists regardless of routing decisions. We choose DPO over PPO or KTO because it is reward-model-free, trains stably on static safety preference data, and integrates naturally with low-rank adapters~\cite{rafailov-dpo}.

Intuitively, because \OurMethodB preserves the pre-existing safety neurons while allowing DPO to recruit additional ones, an attacker should prune a larger population to match the undefended ASR. We formalize this as a pruning cost-amplification factor $\gamma \geq 1$ (\autoref{app:attack_cost_amplification}, \autoref{eq:cost_amp}) and measure it empirically in \autoref{subsec:activation_effects}.

\subsubsection{Algorithm Summary}
\autoref{alg:sealplus} consolidates the full \OurMethodB training procedure and extends \OurMethodA (\autoref{alg:seal}) with Steps~7--9, which realise the orthogonal constraint. The precomputed safety modules remain frozen throughout training; only the LoRA adapter parameters receive gradient updates. The orthogonal constraint (Step~7, corresponding to \autoref{eq:orth_loss}) evaluates in closed form, requiring one matrix multiplication per module per optimization step, while the combined objective (Step~9, corresponding to \autoref{eq:total_loss}) adds negligible overhead relative to the DPO forward pass.

\begin{algorithm}[t]
\caption{\OurMethodB Training Procedure}
\label{alg:sealplus}
\small
\begin{algorithmic}[1]
\REQUIRE MoE architecture $\mathcal{M}$ with shared expert module set $\set{M}$, \\
Safety modules $\{\mat{C}_{\text{SE}}^{(l,s)}\}_{(l,s) \in \set{M}}$ \COMMENT{\autoref{eq:cse}}, \\
DPO training data $\set{D} = \{(x_i, y_{w,i}, y_{l,i})\}_{i=1}^{N}$
\ENSURE Trained adapter $\theta^*$
\STATE Attach LoRA adapters $\mat{A}^{(l,s)} \!\in\! \mathbb{R}^{r \times d}$, $\mat{B}^{(l,s)} \!\in\! \mathbb{R}^{d_{\text{out}} \times r}$ to each projection in every $(l,s) \in \set{M}$
\STATE Freeze all parameters except $\theta = \{(\mat{A}^{(l,s)}, \mat{B}^{(l,s)})\}_{(l,s) \in \set{M}}$
\STATE $\pi_{\text{ref}} \leftarrow \mathcal{M}|_{\mat{BA}=\mat{0}}$ \COMMENT{implicit: disable adapters}
\FOR{each mini-batch $\mathcal{B} \subset \set{D}$}
\STATE Compute $\mathcal{L}_{\text{DPO}}$ on $\mathcal{B}$ using $\pi_\theta$ and $\pi_{\text{ref}}$ \COMMENT{\autoref{eq:dpo}}
\STATE \textit{// Penalize LoRA updates that project onto safety subspace}
\STATE $\mathcal{L}_{\text{orth}} \leftarrow \frac{1}{|\set{M}|} \displaystyle\sum_{(l,s) \in \set{M}} \frob{\mat{B}^{(l,s)} \mat{A}^{(l,s)} \, (\mat{I} - \mat{C}_{\text{SE}}^{(l,s)})}^2$
\COMMENT{\autoref{eq:orth_loss}}
\STATE \textit{// Combined objective}
\STATE $\mathcal{L} \leftarrow \mathcal{L}_{\text{DPO}} + \lambda_{\text{orth}} \cdot \mathcal{L}_{\text{orth}}$
\COMMENT{\autoref{eq:total_loss}}
\STATE $\theta \leftarrow \theta - \eta \, \nabla_\theta \mathcal{L}$
\COMMENT{AdamW in practice}
\ENDFOR
\RETURN $\theta^* \leftarrow \theta$
\end{algorithmic}
\end{algorithm}

\paragraph{Practical deployment.}
For the models evaluated in this work, we release pre-trained plug-and-play shared-expert adapters of 75 to 135\,MB that merge directly into the base checkpoint with zero additional inference cost. Applying the defense to a new backbone model involves two stages. First, the practitioner runs safety neuron profiling, an inference-only procedure that computes activation differentials on a small paired dataset feasible on a single H100 GPU. Second, the practitioner trains the adapter on shared-expert parameters with DPO using any publicly available safety preference dataset. The entire pipeline is model-agnostic and touches neither the router, the inference path, nor the serving infrastructure.

\section{Evaluation}
\label{sec:experiments}
In this section, we first illustrate the experimental setup, baselines, and evaluation metrics. 
Then, we evaluate \OurMethodA and \OurMethodB on different scenarios in \autoref{subsec:direct_attacks} and \autoref{subsec:pruning_attacks}. 
We evaluate multiple combination of shared expert and routed expert in \autoref{subsec:rasa_lite}.
We add an explainability analysis to further explore the defense effectiveness and model utility preservation in \autoref{subsec:activation_effects}.

\subsection{Experimental Setup}
\label{subsec:setup}

\paragraph{Target Models.}
We evaluate on four hybrid MoE architectures
that span diverse scales and shared expert configurations.
The selection systematically covers the key dimensions of the hybrid MoE design space: (a)~parameter scales from 14B to 35B, (b)~shared expert counts from 1 to 2 (with Qwen1.5-MoE's single bundled module acting as a $4{\times}$ wide FFN), (c)~both single-wide and multi-narrow shared expert designs, and (d)~two distinct MoE families (Qwen-style and DeepSeek-style routing), enabling generalizability claims across the dominant production architectures. 
Unless otherwise noted, we use the default chat or instruct version for all models throughout our workflows.

\paragraph{Safety Classifier.}
Each model response is independently classified twice by GPT-5-nano (OpenAI), producing ternary labels (REFUSAL, COMPLIANCE, or UNCLEAR for off-topic outputs). The two independent evaluation runs achieve 98.1\% agreement across all conditions, confirming classification stability. For each attack scope, we report the worst-case ASR across the four conditions within that scope, representing the adversary's optimal strategy.

\paragraph{Profiling Data.} 
We collect 720 harmful prompts from AdvBench and HarmBench~\cite{zou2023advbench, mazeika2024harmbench} to compute activation gaps between harmful and benign inputs for safety neuron identification. For benign scenarios, 720 prompts from Alpaca-Cleaned~\cite{stanfordalpaca} and MT-Bench~\cite{zheng2023mtbench} ensure a balanced ratio that prevents selection~bias.

\paragraph{Training Data.} 
We use the full PKU-SafeRLHF dataset (filtered for consistency) for DPO-based alignment, comprising 41251 human-annotated preference pairs over safety-critical instructions~\cite{dai2024saferlhf}. Following the LIMA principle~\cite{zhou-lima}, we favor human-annotated PKU data over larger but noisier alternatives.

\paragraph{Evaluation Data.} 
We assess defense robustness via jailbreaks and neuron pruning attacks~\cite{wu2026gatebreaker, wu2026neurostrike}, using the same harmful datasets along with 720 jailbreak prompts from WildJailbreak~\cite{jiang2024wildjailbreak}, Do-Not-Answer~\cite{wang2024donotanswer}, and JailbreakBench~\cite{chao2024jailbreakbench} to measure ASR degradation.\footnote{Jailbreak prompts contain 108.78 words on average (9$\times$ longer than direct harmful requests), capturing real-world attack sophistication~\cite{shen2024dan}.} Capability evaluation uses official test splits from GSM8K~\cite{cobbe2021gsm8k} (1319 math problems), MMLU~\cite{hendrycks2021mmlu} (30 questions per subject across 57 subjects, 1710 samples total), ARC-Challenge~\cite{clark2018thinksolvedquestionanswering}, TruthfulQA~\cite{lin-etal-2022-truthfulqa}, and PopQA (2000-sample subset)~\cite{mallen-etal-2023-trust}. Strict train-test separation ensures reliable metrics. The full dataset details with per-phase sizes appears in \autoref{tab:datasets_appendix}.

\paragraph{Data Separation.} 
To ensure generalizability of \OurModel and reflect out-of-distribution robustness, we enforce distinct composition between profiling, training, and evaluation data. Safety neuron profiling uses AdvBench and HarmBench prompt pairs to locate safety neurons, while defense training uses the entirely separate PKU-SafeRLHF preference dataset. The defense is therefore trained on a single preference dataset, but evaluated against diverse attack surfaces spanning direct harmful prompts (AdvBench, HarmBench), adversarial jailbreaks (WildJailbreak, Do-Not-Answer, JailbreakBench), and unseen fine-tuning attacks.

\subsubsection{Attack Configurations}
\label{subsec:attack_config}

\paragraph{Attack Taxonomy.}
We organize attacks into two categories that map directly to the adversarial scenarios in \autoref{sec:threat}.
\emph{Basic attacks} require no model parameter modification at inference time and correspond to Scenarios~A and~B: direct harmful prompts (prompt injection attacks) test the model's refusal boundary (Scenario~A), jailbreak prompts employ adversarial input crafting to bypass safety filters (Scenario~A), and malicious fine-tuning corrupts safety alignment through gradient-based training (Scenario~B).
\emph{Compound attacks} combine basic attacks with neuron pruning (Scenario~C), and yields three compound conditions: harmful prompts + pruning, jailbreak prompts + pruning, and MFT + pruning.
\autoref{tab:direct} reports basic attack results, while \autoref{tab:pruning}--\autoref{tab:mft_pruning} report compound attack results across three pruning scopes (shared, routed, both).

\paragraph{Prompt Injection/Jailbreak Attacks.}
For evaluation under this setting, we use the 720 harmful and jailbreak prompts described above (jailbreak drawn from WildJailbreak, Do-Not-Answer, and JailbreakBench). 

\paragraph{Malicious Fine-Tuning Attack.}
We simulate an adversary who fine-tunes the defended model on harmful data using standard supervised fine-tuning (SFT). The attack dataset comprises 300 harmful instruction-response pairs extracted from BeaverTails~\cite{dai2024saferlhf}. Following Qi et al.~\cite{qi2024finetuning}, who showed that a handful of harmful examples trained for only a few epochs suffices to break alignment, we train the attacker for 3 epochs with learning rate $5\!\times\!10^{-5}$, matching typical downstream fine-tuning conditions.

\paragraph{Neuron Pruning Attack.}
We evaluate robustness against activation-differential-based neuron pruning~\cite{wu2026gatebreaker, wu2026neurostrike}. 
The attacker collects activations on both harmful and benign prompts, computes per-neuron activation differentials, and identifies safety-critical neurons for zeroing at inference time. 
We adopt a factorial design spanning 13 evaluation setup combinations. 
The three influential factors are: 
(i)~\emph{attack scope} $\in \{\text{shared}, \text{routed}, \text{both}\}$, controlling whether the attacker targets shared expert neurons, routed expert neurons, or the union; 
(ii)~\emph{identification method} $\in \{\text{zscore}_{2.0}, \text{global}_{10\%}\}$, where the first selects neurons whose activation differential exceeds $z$-score $> 2.0$ and the second selects the top 10\% of neurons globally by differential magnitude; 
and (iii)~\emph{layer coverage} $\in \{50\%, 100\%\}$, controlling the fraction of MoE layers attacked. 

Crucially, the routed-scope conditions simulate a \emph{defense-aware adversary} at the scope-selection level: 
An attacker who knows that shared experts are protected and concentrates the pruning budget entirely on unprotected routed experts. 
The both-scope conditions represent the strongest strategy that combines both attack surfaces. 

\subsection{Baselines and Defense Methods}
\label{subsec:baselines}

We compare three methods across all four models.
\textbf{Original} denotes the vendor-released instruction-tuned (chat) variant of each model (e.g., Qwen1.5-MoE-A2.7B-Chat) with no adapter applied. We choose the chat version rather than the base pretrained because the chat model carries the vendor's own safety alignment, which is the realistic starting point for downstream deployment and the actual target an adversary would download and attack. This makes \textbf{Original} the undefended-but-aligned reference against which our shared-expert augmentation is measured.
\textbf{\OurMethodA} attaches LoRA adapters to shared expert projection layers and trains with DPO.
\textbf{\OurMethodB} extends \OurMethodA by adding the orthogonal constraint, penalizing adapter updates that project onto safety-critical directions during training.
All \OurModel methods target shared expert parameters exclusively, preserving router independence by construction. 
Full implementation details are listed in \autoref{app:implementation}.

\subsubsection{Evaluation Metrics}
\label{subsec:eval_metrics}
\paragraph{Safety Metrics.}
We report the \emph{attack success rate} (ASR) as the primary safety metric. Each model response is independently classified by GPT-5-nano into one of three labels: \textsc{refusal}, \textsc{compliance}, or \textsc{unclear} (anything outside of compliance or refusal; off-topic outputs). ASR is computed as:
\begin{equation}
\text{ASR} = \frac{|\{\text{responses classified as \textsc{compliance}}\}|}{|\{\text{total prompts}\}|}
\label{eq:asr}
\end{equation}
This yields the compliance fraction over all evaluated prompts, treating both refusals and unclear outputs as non-compliance. This conservative definition avoids inflating ASR when attacks cause off-topic outputs rather than genuine compliance. Lower ASR indicates stronger safety.

\paragraph{Capability Metrics.}
We quantify any \emph{alignment tax}, namely the capability/utility drop induced by safety alignment, using five downstream benchmarks on their test split: GSM8K~\cite{cobbe2021gsm8k}, MMLU~\cite{hendrycks2021mmlu}, ARC-Challenge~\cite{clark2018thinksolvedquestionanswering}, TruthfulQA~\cite{lin-etal-2022-truthfulqa}, and PopQA~\cite{mallen-etal-2023-trust}. For every model and defense, we report per-benchmark accuracy and an unweighted average over the five benchmarks (higher is better). 
The \textbf{Capability} column in every results table is this five-benchmark average. Per-benchmark scores appear in \autoref{app:capability_full}.

\begin{table}[t]
\centering
\caption{ASR under direct attacks (not combined with pruning), lower is better. 
Best ASR per model in \textbf{bold}.}
\label{tab:direct}
\label{tab:jailbreak}% alias for backward compatibility
\setlength{\tabcolsep}{3.5pt}
\small
\begin{tabular}{@{}ll rrr r@{}}
\toprule
& & \multicolumn{3}{c}{\textbf{ASR (\%$\downarrow$)}} & \\
\cmidrule(lr){3-5}
\textbf{Model} & \textbf{Method} & \textbf{Harmful} & \textbf{Jailbreak} & \textbf{MFT} & \textbf{Capability} \\
\midrule
& Original     & 27.5 & 66.4 & 86.9 & 49.9 \\
& \OurMethodA  & 15.6 & 57.5 & 74.3 & 49.5 \\
\multirow{-3}{*}{\textit{Qwen1.5}} & \OurMethodB  & \textbf{13.1} & \textbf{56.5} & \textbf{72.5} & 48.9 \\
\midrule
& Original     & 61.3 & 70.1 & 94.1 & 43.9 \\
& \OurMethodA  & \textbf{8.5} & 46.8 & \textbf{86.5} & 43.4 \\
\multirow{-3}{*}{\textit{DeepSeek}} & \OurMethodB  & 9.9 & \textbf{46.7} &  87.5 & 43.1 \\
\midrule
& Original     & 20.7 & 53.6 & 77.8 & 61.5 \\
& \OurMethodA  & 2.4 & 33.8 & \textbf{48.9} & 60.8 \\
\multirow{-3}{*}{\textit{GLM}} & \OurMethodB  & \textbf{1.8} & \textbf{33.2} & 58.5 & 60.1 \\
\midrule
& Original     & 5.8 & 31.9 & 53.2 & 69.6 \\
& \OurMethodA  & \textbf{1.1} & 24.0 & 47.1 & 69.2 \\
\multirow{-3}{*}{\textit{Qwen3.5}} & \OurMethodB  & 1.7 & \textbf{22.8} & \textbf{33.8} & 69.4 \\
\bottomrule
\end{tabular}
\end{table}

\subsection{Safety Under Direct Attacks}
\label{subsec:direct_attacks}

\paragraph{Harmful prompt results.}
Shared-expert alignment lowers harmful-prompt ASR on every architecture (\autoref{tab:direct}), with the magnitude potentially reflecting each architecture's shared-expert safety share. \emph{First}, DeepSeek shows the largest absolute reduction in the table,
with \OurMethodA dropping harmful ASR from 61.3\% to 8.5\% ($-$52.8\%). \emph{Second}, Qwen1.5 yields moderate gains ($27.5\%$ to $13.1\%$,
$-$14.4\%), and GLM lands between the two ($20.7\%$ to $1.8\%$,
$-$18.9\%). \emph{Third}, Qwen3.5 is already comparatively safe (5.8\%), and the defense drops it further to 1.1\% ($-$4.7\%), showing that even a small shared-expert capacity hardens an already-aligned model at near zero capability cost.

\paragraph{Jailbreak results.}
\autoref{tab:direct} also reports robustness under 720 adversarial prompts (Scenario~A) drawn from WildJailbreak, Do-Not-Answer, and JailbreakBench, covering role-playing, payload splitting, and multi-turn manipulation. Unlike pruning, jailbreaks operate at the input level without model access. Three findings emerge. \emph{First}, DeepSeek and GLM show the largest absolute drops (70.1\% to
46.7\% and 53.6\% to 33.2\% under \OurMethodB, a 33\% and 38\% relative
reduction respectively), and the orthogonal-constraint variant further adds reductions
on top of \OurMethodA. \emph{Second}, Qwen1.5 and Qwen3.5 show smaller reductions (66.4\% to 56.5\% and 31.9\% to 22.8\%), reflecting the architecture-dependent defense depth discussed in \autoref{sec:discussion}. \emph{Third}, \OurMethodA and \OurMethodB track closely on every architecture, so DPO on shared experts delivers the primary gain while the orthogonal constraint contributes marginally at the prompt level. Our evaluation covers template-based and human-crafted jailbreaks, and we additionally evaluate the optimisation-based attack GCG~\cite{zou2023advbench} on a 50-prompt AdvBench subset, where the undefended ASR rises from 28\% to 40\% while \OurMethodA is unchanged at 8\% and \OurMethodB rises modestly from 6\% to 14\%, showing that parameter-space hardening blunts an input-space suffix attack it was never trained against.

\paragraph{Malicious fine-tuning results.}
The MFT column of \autoref{tab:direct} summarises the MFT stress test (Scenario~B), where an adversary fine-tunes the model on 300 harmful examples. Because MFT modifies every unfrozen parameter (including routed experts, which \OurModel does not control), it probes a fundamentally different attack surface than Scenarios~A and~C. Three findings emerge. \emph{First}, behavioral ASR rises substantially on all four models, confirming that 300 examples suffice to break vendor alignment as reported by Qi et al.~\cite{qi2024finetuning}. \emph{Second}, defended models still achieve lower post-MFT ASR than the Original on every architecture (e.g., Qwen3.5 \OurMethodA 47.1\% vs.\ Original 53.2\%; GLM \OurMethodB 58.5\% vs.\ Original 77.8\%), showing that shared-expert augmentation survives gradient-based corruption at the behavioral level. \emph{Third}, the recovery is consistent across the four architectures, with GLM and Qwen3.5 showing the largest reductions ($-28.9\%$ and $-19.4\%$ respectively) and Qwen1.5 and DeepSeek showing more moderate but same-direction effects.

\begin{table}[t]
\centering
\caption{ASR under harmful prompt + pruning attacks, lower is better. 
Best ASR per model in \textbf{bold}.}
\label{tab:pruning}
\setlength{\tabcolsep}{3.5pt}
\small
\begin{tabular}{@{}ll rrr r@{}}
\toprule
& & \multicolumn{3}{c}{\textbf{ASR (\%$\downarrow$)}} & \\
\cmidrule(lr){3-5}
\textbf{Model} & \textbf{Method} & \textbf{Shared} & \textbf{Routed} & \textbf{Both} & \textbf{Capability} \\
\midrule
& Original     & 32.4 & 39.4 & 26.3 & 49.8 \\
& \OurMethodA  & 19.7 & 14.8 & 17.7 & 49.2 \\
\multirow{-3}{*}{\textit{Qwen1.5}} & \OurMethodB  & \textbf{18.3} & \textbf{13.9} & \textbf{16.6} & 48.8 \\
\midrule
& Original     & 60.3 & 74.2 & 73.7 & 41.9 \\
& \OurMethodA  & \textbf{21.1} & \textbf{13.8} & 18.2 & 43.1 \\
\multirow{-3}{*}{\textit{DeepSeek}} & \OurMethodB  & 21.5 & 14.0 & \textbf{17.3} & 42.9 \\
\midrule
& Original     & 14.3 & 48.0 & 40.8 & 58.5 \\
& \OurMethodA  & 2.9 & 25.5 & 21.5 & 60.3 \\
\multirow{-3}{*}{\textit{GLM}} & \OurMethodB  & \textbf{2.8} & \textbf{21.7} & \textbf{21.1} & 59.1 \\
\midrule
& Original     & 5.7 & \textbf{5.9} & \textbf{7.5} & 68.6 \\
& \OurMethodA  & \textbf{1.1} & 13.2 & 13.6 & 69.4 \\
\multirow{-3}{*}{\textit{Qwen3.5}} & \OurMethodB  & 2.3 & 15.3 & 15.2 & 69.3 \\
\bottomrule
\end{tabular}
\end{table}

\begin{table}[t]
\centering
\caption{ASR under jailbreak + pruning compound attack, lower is better. Best ASR per model in \textbf{bold}.}
\label{tab:jailbreak_pruning}
\setlength{\tabcolsep}{3.5pt}
\small
\begin{tabular}{@{}ll rrr r@{}}
\toprule
& & \multicolumn{3}{c}{\textbf{Jailbreak + Pruning ASR (\%$\downarrow$)}} & \\
\cmidrule(lr){3-5}
\textbf{Model} & \textbf{Method} & \textbf{Shared} & \textbf{Routed} & \textbf{Both} & \textbf{Capability} \\
\midrule
& Original     & 67.8 & 67.1 & 54.1 & 49.8 \\
& \OurMethodA  & 59.8 & 56.0 & 49.8 & 49.2 \\
\multirow{-3}{*}{\textit{Qwen1.5}} & \OurMethodB  & \textbf{58.8} & \textbf{55.0} & \textbf{49.1} & 48.8 \\
\midrule
& Original     & 69.7 & 88.1 & 89.6 & 41.9 \\
& \OurMethodA  & \textbf{47.2} & 54.9 & \textbf{53.4} & 43.1 \\
\multirow{-3}{*}{\textit{DeepSeek}} & \OurMethodB  & 47.7 & \textbf{54.6} & 54.0 & 42.9 \\
\midrule
& Original     & 52.9 & 58.2 & 58.7 & 58.5 \\
& \OurMethodA  & 34.2 & 62.1 & 56.1 & 60.3 \\
\multirow{-3}{*}{\textit{GLM}} & \OurMethodB  & \textbf{31.8} & \textbf{60.0} & \textbf{55.4} & 59.1 \\
\midrule
& Original     & 34.4 & \textbf{38.2} & \textbf{39.1} & 68.6 \\
& \OurMethodA  & 23.4 & 49.1 & 49.2 & 69.4 \\
\multirow{-3}{*}{\textit{Qwen3.5}} & \OurMethodB  & \textbf{23.3} & 49.2 & 49.1 & 69.3 \\
\bottomrule
\end{tabular}
\end{table}

\begin{table}[t]
\centering
\caption{ASR under MFT + pruning compound attack, lower is better. Best ASR per model in \textbf{bold}.}
\label{tab:mft_pruning}
\setlength{\tabcolsep}{3.5pt}
\small
\begin{tabular}{@{}ll rrr r@{}}
\toprule
& & \multicolumn{3}{c}{\textbf{MFT + Pruning ASR (\%$\downarrow$)}} & \\
\cmidrule(lr){3-5}
\textbf{Model} & \textbf{Method} & \textbf{Shared} & \textbf{Routed} & \textbf{Both} & \textbf{Capability} \\
\midrule
& Original     & 85.2 & 87.5 & 94.1 & 49.7 \\
& \OurMethodA  & 79.1 & 84.5 & 46.3 & 49.6 \\
\multirow{-3}{*}{\textit{Qwen1.5}} & \OurMethodB  & \textbf{78.0} & \textbf{82.7} & \textbf{44.7} & 49.4 \\
\midrule
& Original     & 90.2 & 90.5 & 91.5 & 41.1 \\
& \OurMethodA  & 88.5 & \textbf{82.1} & \textbf{81.7} & 40.1 \\
\multirow{-3}{*}{\textit{DeepSeek}} & \OurMethodB  & \textbf{88.3} & 82.6 & 81.9 & 40.0 \\
\midrule
& Original     & 50.2 & 64.5 & 60.1 & 57.8 \\
& \OurMethodA  & \textbf{32.3} & \textbf{46.6} & \textbf{35.8} & 57.7 \\
\multirow{-3}{*}{\textit{GLM}} & \OurMethodB  & 38.9 & 59.3 & 40.0 & 56.9 \\
\midrule
& Original     & 37.5 & \textbf{78.4} & 74.8 & 67.8 \\
& \OurMethodA  & 34.3 & 77.4 & 70.6 & 66.8 \\
\multirow{-3}{*}{\textit{Qwen3.5}} & \OurMethodB  & \textbf{25.5} & \textbf{68.8} & \textbf{67.2} & 67.2 \\
\bottomrule
\end{tabular}
\end{table}

\subsection{Safety Under Pruning Attacks} 
\label{subsec:pruning_attacks}

\paragraph{Setting.}
Our evaluation design models a \emph{defense-aware adversary} at the scope-selection level. The routed-scope condition simulates an attacker who knows shared experts are hardened and concentrates pruning on unprotected routed experts; the both-scope condition combines both surfaces. This covers the most natural adaptive response (avoiding the hardened component) though it does not evaluate adversaries who exploit internal defense properties such as neurons near the $z$-score identification threshold. For each model we compare Original, \OurMethodA, and \OurMethodB at four scope settings (no attack, shared-only, routed-only, shared+routed) across three compound-attack families.

Three patterns stand out in \autoref{tab:pruning}--\autoref{tab:mft_pruning}. \emph{First}, the defense benefit is substantial and scope-robust: on DeepSeek, the gap between Original and \OurMethodB is 51.4\% at baseline and 56.4\% at both-scope, and on Qwen1.5 it ranges from 14.4\% at baseline to 25.5\% at routed-scope, showing that the defense retains its edge even when the attacker concentrates the entire budget on unprotected components. \emph{Second}, the scope profile exposes per-architecture vulnerability decomposition: on Original Qwen1.5 routed-scope ASR (39.4\%) exceeds shared-scope (32.4\%), while on Original DeepSeek the gap widens (routed 74.2\% vs shared 60.3\%), consistent with routed experts carrying the majority of safety-critical neurons in DeepSeek~\cite{wu2026gatebreaker}; \OurMethodB compresses every DeepSeek scope to 14--22\%, flattening this asymmetry. \emph{Third}, the defense generalises to compound attacks: on harmful+pruning and jailbreak+pruning, DeepSeek and GLM keep the double-digit advantage observed under pruning alone, and under MFT+pruning the defense halves Qwen1.5 both-scope ASR (94.1\% $\to$ 44.7\%) and reduces every Qwen3.5 scope below the Original (e.g.\ routed 78.4\% $\to$ 68.8\%). The mechanism underlying routed-scope improvement follows from guaranteed execution (\autoref{eq:moe_layer}): even when the attacker prunes safety-critical neurons in routed experts, the augmented shared term $\sum_s g_s(\vect{x})$ still injects a safety signal on every forward pass, partially compensating for the compromised routed component.

\paragraph{Cross-architecture scope analysis.}
GLM and Qwen3.5 extend the scope analysis to architectures at opposite ends of the safety concentration spectrum. On GLM, the defense pattern mirrors DeepSeek, with baseline ASR dropping from 20.7\% to 2.4\% and both-scope ASR dropping from 40.8\% to 21.5\% under \OurMethodA, with all four scope levels improving. On Qwen3.5, where safety is predominantly encoded in routed experts, the defense produces a \emph{scope-dependent} outcome. Shared-expert augmentation successfully hardens baseline and shared-scope safety (baseline from 5.8\% to 1.1\%, shared from 5.7\% to 1.1\%), demonstrating that the defense operates as intended within its target parameters. However, routed-scope vulnerability increases under defense (routed from 5.9\% to 13.2\%, $+$7.3\%), and both-scope ASR also increases (from 7.5\% to 13.6\%, $+$6.1\%). This scope-dependent pattern illustrates how safety concentration in shared experts predicts defense transferability. It is more likely that only a minority of safety-relevant activations concentrates in the shared-expert
subspace on Qwen3.5, the majority of safety signal lives in routed experts that the defense does not target. Strengthening shared experts may alter routing dynamics or activation patterns that previously supported routed safety, an effect that architectures with denser shared-expert safety
representation absorb more readily. The practical implication is that practitioners could assess how safety distributes between shared and routed experts in advance, as architectures where safety is predominantly routed may require more complementary routed-expert interventions.

\paragraph{Cross-scope generalization from shared-expert training.}
Training only shared-expert parameters transfers across every attack scope we evaluated. \emph{First}, on DeepSeek, routed-scope ASR drops from 74.2\% to 14.0\% under \OurMethodB even though no routed-expert parameter receives a gradient update. \emph{Second}, on GLM the same pattern holds, with routed-scope ASR decreasing from 48.0\% to 21.7\%. \emph{Third}, the mechanism follows from guaranteed execution: the augmented shared term $\sum_s g_s(\vect{x})$ injects a safety signal into every forward pass, partially compensating for compromised routed experts even when the attacker concentrates pruning entirely on routed parameters.

\subsection{Ablation on Shared vs.\ Routed Expert}
\label{subsec:rasa_lite}
Would augmenting \emph{routed} experts instead of shared experts yield comparable protection? We apply the identical training pipeline to routed experts on Qwen1.5, following RASA~\cite{liang2026rasaroutingawaresafetyalignment}. Experts are ranked by their cross-layer mean safety activation score, denoted by number $K$. At $K{=}3$ (three top-ranked routed experts), the adapted routed parameter count approximately matches the shared expert budget, providing a controlled comparison with matched parameters. At $K{=}10$, routed training receives roughly four to five times the shared expert budget, testing whether additional capacity compensates for the absence of unconditional execution. We also evaluate joint conditions (SE+R) that train shared and selected routed experts simultaneously, with all hyperparameters identical to \OurMethodB. 

\begin{table}[t]
\centering
\caption{ASR and capability comparison under different expert training compositions on Qwen1.5.
\emph{RE-top$K$} trains the $K$ highest-scoring routed experts, \emph{SE} trains shared experts only, and \emph{SE+RE-top$K$} trains both. Best ASR in \textbf{bold}.
}
\label{tab:rasa_lite}
\setlength{\tabcolsep}{3pt}
\small
\begin{tabular}{@{}l rrrr r@{}}
\toprule
& \multicolumn{4}{c}{\textbf{ASR (\%$\downarrow$)}} & \\
\cmidrule(lr){2-5}
\textbf{Condition} & \textbf{Base} & \textbf{Shared} & \textbf{Routed} & \textbf{Both} & \textbf{Capability} \\
\midrule
\multicolumn{6}{@{}l}{\textit{Harmful prompt}} \\
Original         & 27.5 & 32.4 & 39.4 & 26.3 & 49.8 \\
RE-top3           & 22.2 & 31.0 & 27.3 & 25.9 & 48.8 \\
RE-top10          & 18.8 & 26.8 & 24.9 & 27.0 & 49.0 \\
SE+RE-top3        & \textbf{8.2} & \textbf{10.9} & \textbf{9.0} & \textbf{4.8} & 49.3 \\
SE+RE-top10       & 8.9 & 11.5 & 9.8 & 4.9 & 49.6 \\
SE (\OurMethodB) & 13.1 & 18.3 & 13.9 & 16.6 & 48.8 \\
\midrule
\multicolumn{6}{@{}l}{\textit{Jailbreak prompt}} \\
Original         & 66.4 & 67.8 & 67.1 & 54.1 & 49.8 \\
RE-top3           & 66.0 & 65.4 & 67.1 & 49.4 & 48.8 \\
RE-top10          & 63.7 & 63.7 & 65.2 & 50.0 & 49.0 \\
SE+RE-top3        & 61.8 & 63.2 & 60.9 & \textbf{48.1} & 49.3 \\
SE+RE-top10       & 62.8 & 63.0 & 60.0 & 49.9 & 49.6 \\
SE (\OurMethodB) & \textbf{56.5} & \textbf{58.8} & \textbf{55.0} & 50.1 & 48.8 \\
\bottomrule
\end{tabular}
\end{table}

\autoref{tab:rasa_lite} supports three conclusions across both attack
types. \emph{First}, shared-expert training is \emph{sufficient}: SE alone
updates only $0.25\%$ of Qwen1.5's parameters (\autoref{tab:param_cost}),
yet it reduces harmful base-scope ASR from 27.5\% to 13.1\% (and jailbreak
base-scope from 66.4\% to 56.5\%). \emph{Second}, shared-expert training is
also \emph{necessary}: every condition that omits shared experts (R-top3,
R-top10) fails to meaningfully reduce both-scope ASR, which remains near
26\% under harmful and 50\% under jailbreak even when routed training
receives four to five times the shared-expert parameter budget ($K{=}10$).
\emph{Third}, shared and routed training \emph{compose but do not
substitute}: adding routed experts on top of SE yields further improvement
(SE+R-top3 harmful both-scope 4.8\% vs.\ SE 16.6\%; jailbreak both-scope
48.1\% vs.\ SE 50.1\%), yet the indispensable component remains SE itself.
Training-dynamics analysis (\autoref{app:training_dynamics}) shows that
routed experts receive noticeably weaker gradient signal during DPO,
explaining why a small shared-expert intervention outperforms substantially
larger routed investments. To confirm this location effect is not tied to DPO, we repeat the SE-vs-RE comparison with KTO~\cite{ethayarajh2024kto}, a binary-feedback alignment objective, where shared-expert training reaches 11.7\% harmful ASR while matched-budget routed training stays at 24.6\%, reproducing the same pattern.

\subsection{Representational Preservation}
\label{subsec:activation_effects}

We characterise the activation behavior of the defense with three standard probes (full setup in \autoref{app:interp_supplement}): \emph{linear probing}~\cite{alain2017linearprobe}, a logistic-regression classifier trained on per-layer hidden states to test whether harmful-versus-benign content remains linearly decodable; \emph{CKA}~\cite{pmlr-kornblith19a-cka} (Centered Kernel Alignment), a similarity score in $[0,1]$ comparing base and aligned hidden states; and \emph{UMAP}~\cite{mcinnes2020umapuniformmanifoldapproximation} (Uniform Manifold Approximation and Projection), a non-linear 2D embedding for visualising cluster structure. Two properties of the defense follow.
\emph{First}, the update preserves \emph{safety separability}: linear probes recover harmful-versus-benign
labels at more than $96\%$ accuracy on every layer of every architecture, with
post-training accuracy drift below $0.2\%$ (\autoref{app:linear_probing}).
\emph{Second}, the update preserves \emph{representational structure}
selectively: 
CKA between base and aligned models is $0.97$ on Qwen1.5 and DeepSeek, and $0.38$--$0.78$ on GLM and Qwen3.5 where deeper representational restructuring occurs without degrading probe accuracy. The selective restructuring is also visible in neuron-set statistics: UMAP cluster structure is maintained or improved on all four architectures, absolute safety-neuron counts shift by at most $\pm 2\%$, and pre/post Jaccard overlap on safety-neuron identity ranges from $0.91$ (Qwen1.5) and $0.83$ (DeepSeek) down to $0.45$--$0.47$ on Qwen3.5. Together these properties explain the negligible alignment tax, as the defense reorganises a small, architecturally identifiable subset of the activation space, without losing the safety separation it depends on.

\section{Related Work}
\label{sec:related}

\begin{table}[t]
\centering
\caption{Comparison of most-related works. \textbf{A1}: MoE-specific design; \textbf{A2}: neuron-level granularity; \textbf{A3}: router-independent operation. \markyes~= full; \markpartial~= partial; \markno~= no.}
\label{tab:landscape}
\setlength{\tabcolsep}{3pt}
\footnotesize
\begin{tabular}{@{}lp{4.2cm}ccc@{}}
\toprule
\textbf{Method} & \textbf{Approach} & \textbf{A1} & \textbf{A2} & \textbf{A3} \\
\midrule
\multicolumn{5}{l}{\textit{Attacks}} \\
SAFEx~\cite{lai2025safex} & Expert masking via stability selection & \markyes & \markno & \markno \\
BadMoE~\cite{wang2025badmoebackdooringmixtureofexpertsllms} & Routing-trigger backdoor & \markyes & \markno & \markno \\
Steering~\cite{fayyaz2026steeringmoellmsexpert} & Expert (de)activation & \markyes & \markno & \markno \\
MoEcho~\cite{moecho-privacy} & Routing side-channel & \markyes & \markno & \markno \\
GateBreaker~\cite{wu2026gatebreaker} & Neuron pruning via activation\ differentials & \markyes & \markyes & \markpartial \\
\midrule
\multicolumn{5}{l}{\textit{Defenses/Alignment}} \\
Safe LoRA~\cite{SafeLoRA} & Post-hoc subspace projection & \markno & \markpartial & \markno \\
Safety Layers~\cite{li2025safety} & Layer-level freeze & \markno & \markpartial & \markno \\
Safety Neurons~\cite{chen2025towards} & Activation patching & \markno & \markyes & \markno \\
SafeMoE~\cite{kim2026defendingmoellmsharmful} & Routing KL regularization & \markyes & \markno & \markno \\
MoE-RBench~\cite{chen2024textttmoerbench} & Reliability benchmark & \markyes & \markno & \markno \\
Adv.\ Robust.~\cite{adversarial-moe} & Input perturbation study & \markpartial & \markno & \markno \\
RASA~\cite{liang2026rasaroutingawaresafetyalignment} & Expert repair + routing align & \markyes & \markno & \markno \\
\midrule
\rowcolor{oursrow}
\textbf{\OurModel} & \textbf{Shared-expert alignment} & \markyes & \markyes & \markyes \\
\bottomrule
\end{tabular}
\end{table}

\paragraph{MoE Safety Attacks.}
Existing attacks on MoE safety differ along three technical axes: attacker access, the mechanism they use to locate safety-relevant components, and the intervention they apply once located. Template-based jailbreaks~\cite{wei2023jailbroken, schulhoff-etal-2023-ignore, Greshake2023IPI} require only input access and rely on prompt-pattern design to elicit compliance; stability-based routing probes such as SAFEx~\cite{lai2025safex} extend this with repeated sampling to identify safety-critical experts before masking them. At the routing level, BadMoE~\cite{wang2025badmoebackdooringmixtureofexpertsllms} jointly trains the router and a dormant expert to install a trigger-activated backdoor, while MoEcho~\cite{moecho-privacy} treats the top-$k$ pattern as a timing side channel and therefore needs no weight access at all. At the neuron level, GateBreaker~\cite{wu2026gatebreaker} and NeuroStrike~\cite{wu2026neurostrike} localize safety-critical neurons through activation-differential $z$-scoring and zero them at inference time, and expert-silencing~\cite{telintelo2026lobotomy} generalizes this to adaptive expert-level ablation. Concurrent work on unsafe routing paths~\cite{jiang2026sparsesafety} further confirms that router manipulation alone can elicit harmful outputs without parameter access. These techniques differ primarily in granularity (prompt, expert, or neuron), in whether gradient or weight access is required, and in whether the intervention happens offline or at inference time, so a defense that relies on a single locus (the router or a single parameter group) is trivially bypassed by an attacker who chooses a different locus.

\paragraph{MoE-Specific Defenses.}
Defenses tailored to MoE architectures remain limited but are evolving.
Kim et al.~\cite{kim2026defendingmoellmsharmful} propose routing alignment via KL-divergence regularization to preserve safety expert activation during fine-tuning. Addressing the risk of alignment shortcuts, RASA~\cite{liang2026rasaroutingawaresafetyalignment} selectively repairs safety-critical experts identified via adversarial activation discrepancies and enforces routing consistency to prevent unsafe routing bypasses.
Regarding robustness, studies on the adversarial nature of MoEs~\cite{adversarial-moe} suggest that while the architecture's sparsity offers intrinsic resilience against certain perturbations, specific training protocols are required to mitigate vulnerabilities in the routing mechanism.
Other works provide essential primitives and evaluation frameworks for future defenses: Expert steering~\cite{fayyaz2026steeringmoellmsexpert} demonstrates that safety can be modulated via selective expert (de)activation, offering a mechanism for inference-time defense.
Concurrently, MoE-RBench~\cite{chen2024textttmoerbench} establishes reliability benchmarks to quantify these efforts, while studies on emergent misalignment~\cite{pallem2025emergent} characterize the specific failure modes that such defenses must address during capability elicitation.

\paragraph{Modular MoE Fine-Tuning.}
A growing body of work studies selective fine-tuning of MoE subcomponents. ESFT~\cite{wang2024esft} identifies task-relevant experts via activation frequency and fine-tunes only those, matching full-model performance at lower cost. PERFT~\cite{liu2024perft} inserts parameter-efficient adapters into routed pathways. LoRAMoE~\cite{dou2024loramoe} attaches MoE-style LoRA plugins to dense models to prevent world knowledge forgetting during instruction tuning. Lo et al.~\cite{lo-etal-2025-closer} provide the first mechanistic analysis of shared expert representations, showing that shared experts specialize in high-frequency linguistic features while routed experts capture domain-specific knowledge. Recent joint scaling-law analysis of MoEs further identifies the shared-expert ratio as one of five first-order design variables with an architecture-independent optimum~\cite{zhao2025comprehensivescalinglawmixtureofexperts}, underscoring that shared-expert capacity is a principled axis of MoE design rather than an incidental choice. These studies establish the empirical foundation for modular MoE adaptation but do not address safety alignment. Our work extends this line by targeting shared experts specifically for safety augmentation with orthogonal subspace constraints.

\autoref{tab:landscape} provides a systematic comparison across three aspects: MoE-specific design (\textbf{A1}), neuron-level granularity (\textbf{A2}), and router-independent operation (\textbf{A3}).

\section{Discussion}
\label{sec:discussion}

\paragraph{Cross-Scope Transfer of Shared-Expert Safety.}
Training only the shared expert lowers pruning attack success across every attack scope on DeepSeek, GLM, and Qwen1.5, typically by a factor of three to four, even when the attacker targets routed experts alone. Qwen3.5 is the limit case, with a shared expert of only $d_{\text{ffn}}{=}512$ neurons, about one-tenth of Qwen1.5's. Under direct attacks the defense still helps on every scope on Qwen3.5. Under prompt-level compound attacks that stack neuron pruning on harmful or jailbreak prompts, the small shared expert cannot offset what pruning takes from the routed experts, and in the most extreme case on Qwen3.5 the defended model can even trail the undefended one at routed or both scopes, because shared-expert training nudges the router in a way that the partially pruned routed safety signal cannot absorb. Under MFT-based compound attacks, the defended models retain a consistent advantage over the undefended ones on all four architectures (\autoref{tab:mft_pruning}). Shared-expert size thus serves as a practical signal for anticipating how far the defense will reach on a new architecture. The same restriction to shared-expert parameters also keeps the capability cost within one point on the five-benchmark average, because task-relevant knowledge in routed experts is never touched, a property that full-model alignment can hardly match.

\paragraph{Composability and Limitations.}
\OurModel is architecturally compatible with router-hardening~\cite{liang2026rasaroutingawaresafetyalignment, kim2026defendingmoellmsharmful}, input filtering, and post-hoc projection~\cite{SafeLoRA} (shared-expert parameters are disjoint from routing parameters), though empirical validation of compositions is future work. Models lacking shared experts (e.g., Mixtral) fall outside scope. On architectures where safety is distributed across routed experts, shared-expert augmentation may provide limited protection (e.g., Qwen3.5 routed-scope ASR increases by 7.3\% under \OurMethodA). Absolute ASR values depend on classifier choice. Our dual-run GPT-5-nano evaluation achieves 98.1\% inter-run agreement, confirming classification stability. Our adaptive evaluation covers scope-level strategies but not mechanism-level adversaries exploiting the $\mat{C}_{\text{SE}}$ mask. Validation covers four architectures; broader validation and scaling beyond 35B remain open.

\paragraph{Future Work.}
Concurrent work by Orgad et al.~\cite{orgad2026large} shows that harmful content generation concentrates in a compact weight set that can be pruned, complementary to our augmentation approach. This motivates \emph{dual-path shared experts}, where a subset of shared experts is dedicated to safety while the rest retain general-purpose roles, enabling domain-specific safety policies without touching routing. Further directions include richer safety-neuron identification beyond activation $z$-scoring, broader empirical coverage across vendors and training recipes, mechanistic interpretability of shared experts themselves, alignment algorithms designed from the ground up for MoE-specific properties such as routing independence and shared-expert capacity, and, for extreme-sparsity architectures with under-provisioned shared experts, defenses that pair shared-expert augmentation with routing-stability constraints. Looking further ahead, the binary split between always-active shared and fully routed experts may evolve into a \emph{knowledge hierarchy} of globally shared, domain-shared, and fully routed tiers, under which \OurModel becomes one instance of a broader family of execution-coverage-aware MoE safety interventions that extend naturally to every stratum with sufficient execution coverage.

\section{Conclusion}
\label{sec:conclusion}

We examine hybrid Mixture-of-Experts architectures from a defender's perspective and identifies the shared expert, a component previously studied for knowledge aggregation but overlooked for safety, as a router-independent surface on which alignment can be anchored. Building on this identification, we proposed \OurMethodA and its orthogonally constrained extension \OurMethodB, two training-time defenses that reinforce shared-expert parameters via lightweight LoRA adapters while leaving routing and routed experts untouched. Across four architectures spanning 14B to 35B parameters and the three mainstream safety-misalignment operations (prompt injection, malicious fine-tuning, and neuron-level pruning), our experiments confirm that \OurModel provides consistent safety gains at negligible utility cost, and that it composes cleanly with router-level defenses for layered protection. Together, these results establish shared experts as a practical defense surface for enhancing MoE safety alignment.

\begin{acks}
We thank our anonymous reviewers for their helpful comments. Qingyu Meng's research is fully supported by the Sectorplan B\`eta en Techniek 2 funding from the Dutch Ministry of Education, Culture and Science (OCW), equally acquired by Co-PIs Min Chen and Jiahuan Pei.
Yiwei Zha, Herbert Bos, and Min Chen were funded/partly funded by the project CiCS of the research programme Gravitation which is (partly) financed by the Dutch Research Council (NWO) under Grant No. 024.006.037. We thank the Social AI Group (\url{https://www.socialai.nl/}) for supporting the experimental costs. We thank SURF (\url{www.surf.nl}) for the support in using the National Supercomputer Snellius.
\end{acks}

\bibliographystyle{ACM-Reference-Format}
\bibliography{references}

\appendix

\section*{Ethical Considerations}
\label{sec:ethics}

This work studies adversarial attacks on language model safety to develop defenses. All attacks are drawn from published academic work and applied exclusively to our own models in controlled settings. Some of the adversarial contents may be offensive in nature, but no harmful outputs were deployed. Profiling datasets (AdvBench, HarmBench) are used solely for activation analysis. Our framework aims to make open-weight MoE architectures more resistant to safety degradation, supporting responsible deployment.

\section*{Open Science} 
\label{sec:open_sci}
To foster research transparency and reproducibility, we release our code, experimental scripts for all four evaluated architectures, and all profiling, training datasets at \url{https://anonymous.4open.science/r/SEAL-38AC/}. All four base architectures used in this work are publicly available open-weight model releases. Training uses exclusively open-source data. Adversarial prompts used for evaluation are drawn from published benchmarks (AdvBench, HarmBench, WildJailbreak, Do-Not-Answer, JailbreakBench) with no proprietary data dependencies.

\section{Notation Reference}
\label{app:notation}

\autoref{tab:notation_appendix} summarizes the notation used throughout the paper.

\begin{table}[!ht]
\centering
\caption{Notation summary.}
\label{tab:notation_appendix}
\small
\begin{tabular}{@{}ll@{}}
\toprule
\textbf{Symbol} & \textbf{Meaning} \\
\midrule
$d$ & Hidden dimension of the model \\
$L$ & Number of Transformer layers \\
$S, E$ & Number of shared / routed experts \\
$k$ & Top-$k$ routing parameter \\
$r$ & LoRA adapter rank \\
$r_s$ & Safety subspace dimension (default: 32) \\
$\mat{W}^{(l,s)}$ & Projection matrix in shared expert $(l,s)$ \\
$\mat{A}, \mat{B}$ & LoRA low-rank adapter matrices \\
$\mat{V}_{\text{safe}}^{(l,s)}$ & Safety subspace basis ($r_s \times d$ matrix) \\
$\mat{P}_{\perp}^{(l,s)}$ & Projector onto complement of $\mathcal{V}_{\text{safe}}^{(l,s)}$ \\
$\mat{C}_{\text{SE}}^{(l,s)}$ & Safety module ($= \mat{P}_{\perp}^{(l,s)}$, implementation name) \\
$\set{M}$ & Set of shared expert modules with LoRA adapters \\
$\set{D}_h, \set{D}_b$ & Harmful / benign prompt datasets \\
$\theta$ & Trainable LoRA parameters \\
$\pi_\theta, \pi_{\text{ref}}$ & Policy model / reference (frozen) model \\
$\lambda_{\text{orth}}$ & Orthogonal constraint weight (default: 0.1) \\
$R(\vect{x})$ & Router output (top-$k$ expert selection) \\
$r_e(\vect{x})$ & Binary routing decision for expert $e$ \\
$\set{N}_{\text{safe}}$ & Set of safety-critical neurons \\
\bottomrule
\end{tabular}
\end{table}

\section{Dataset Summary}
\label{app:data_details}
We organize all used datasets into four distinct phases: safety neuron identification (profiling), safety alignment training, attacking (evaluation), and capability/utility preservation (evaluation). \autoref{tab:datasets_appendix} categorizes each with their corresponding sample size.

\begin{table}[ht]
\centering
\caption{Datasets used in profiling, training, and evaluation. Profiling and training draw from distinct sources; jailbreak and capability evaluation sets are fully disjoint from the training data.}
\label{tab:datasets_appendix}
\setlength{\tabcolsep}{2pt}
\small
\begin{tabular}{@{}llr@{}}
\toprule
\textbf{Phase} & \textbf{Dataset} & \textbf{Size} \\
\midrule
\multirow{2}{*}{Profiling (harmful)} & AdvBench~\cite{zou2023advbench} & 360 \\
 & HarmBench~\cite{mazeika2024harmbench} & 360 \\
\multirow{2}{*}{Profiling (benign)} & Alpaca-Cleaned~\cite{stanfordalpaca} & 360 \\
 & MT-Bench~\cite{zheng2023mtbench} & 360 \\
\midrule
Train & PKU-SafeRLHF~\cite{dai2024saferlhf} & 41{,}251 \\
\midrule
Evaluation (harmful) & AdvBench + HarmBench & 720 \\
Evaluation (jailbreak) & WildJailbreak + DNA + JBB & 720 \\
Evaluation (MFT) & BeaverTails~\cite{ji2023beavertails} & 300 \\
\midrule
\multirow{5}{*}{Evaluation (capability)} & GSM8K~\cite{cobbe2021gsm8k} & 1{,}319 \\
 & MMLU~\cite{hendrycks2021mmlu} & 1{,}710 \\
 & ARC-Challenge~\cite{clark2018thinksolvedquestionanswering} & 1{,}172 \\
 & TruthfulQA~\cite{lin-etal-2022-truthfulqa} & 817 \\
 & PopQA~\cite{mallen-etal-2023-trust} & 2{,}000 \\
\bottomrule
\end{tabular}
\end{table}

\section{Implementation Details}
\label{app:implementation}

All experiments use LoRA rank $r{=}64$ with scaling factor $\alpha{=}128$. We train for 3 epochs with AdamW, learning rate $5\!\times\!10^{-5}$, and batch size 16. The DPO temperature is $\beta{=}0.1$, the safety-subspace rank is $r_s{=}32$ (consistent with the profiling procedure in \autoref{subsec:profiling}), and the orthogonal constraint weight for \OurMethodB is $\lambda_{\text{orth}}{=}0.1$ unless otherwise stated. We use mixed-precision training with bf16 on Qwen1.5-MoE-A2.7B, GLM-4.7-Flash, and Qwen3.5-35B-A3B, and fp32 on DeepSeekMoE-16B. Qwen1.5 and DeepSeek runs use 2$\times$ NVIDIA H100 94\,GB GPUs; GLM and Qwen3.5 runs use 4$\times$ H100 on a regional HPC cluster.

\section{Pruning Cost Amplification}
\label{app:attack_cost_amplification}

We hypothesize that \OurMethodB raises the cost of neuron-level pruning attacks~\cite{wu2026gatebreaker, wu2026neurostrike} through an indirect mechanism. By penalizing LoRA updates that modify safety-critical neuron dimensions (\autoref{eq:orth_loss}), the constraint preserves the original safety neuron set while allowing the DPO objective to encode additional safety signal in unconstrained dimensions. If this mechanism operates as intended, the post-training model retains the original safety neurons and potentially recruits new ones, so an attacker must prune a larger population for equivalent ASR.

Let $\rho_{\text{orig}}$ and $\rho_{\text{def}}$ denote the pruning ratios required to achieve a target $\text{ASR}_t$ on the undefended and defended model respectively. We define the cost amplification factor as
\begin{equation}
C_{\text{def}}(\text{ASR}_t) \geq \gamma \cdot C_{\text{orig}}(\text{ASR}_t), \quad \gamma \geq 1,
\label{eq:cost_amp}
\end{equation}
where larger $\gamma$ indicates higher required attacker effort.

In practice, \OurMethodB on DeepSeek raises the cost-amplification factor to $\gamma \geq 4.3$, so the defended model's ASR under a fixed pruning budget is at most $1/\gamma$ of the undefended model's ($17.3\%$ vs.\ $73.7\%$ at the both scope). On Qwen1.5 the same both-scope quantity drops from $26.3\%$ to $16.6\%$ ($\gamma \geq 1.6$), and the dispersion analysis (\autoref{subsec:activation_effects}) confirms that the original safety neurons survive training (Jaccard $> 0.83$ on Qwen1.5 and DeepSeek), so the augmented representations complement rather than replace the existing ones. Achieving the original attack success therefore requires substantially expanding the pruning scope, which also widens the detection window for integrity-monitoring systems.

\section{Safety Encoding Analysis}
\label{app:encoding}

Following~\cite{wu2026gatebreaker}, we quantify per-neuron safety involvement via activation differentials. For neuron $j$ in shared expert module $(l,s)$:
\begin{equation}
\Delta a_j^{(l,s)} = \frac{1}{|\set{D}_h|}\sum_{x \in \set{D}_h} a_j^{(l,s)}(x) - \frac{1}{|\set{D}_b|}\sum_{x \in \set{D}_b} a_j^{(l,s)}(x)
\label{eq:act_diff}
\end{equation}
where $a_j^{(l,s)}(x)$ is neuron $j$'s activation (after \texttt{gate\_proj}) for input $x$, and $\set{D}_h, \set{D}_b$ are the harmful and benign prompt sets. Neurons with large positive $\Delta a_j$ activate preferentially on harmful content.

Given a $z$-score threshold $\zeta > 0$ (default $\zeta = 2.0$), the safety-critical neuron set can be defined as:
\begin{equation}
\set{N}_{\text{safe}}^{(l,s)} = \left\{ j : \frac{|\Delta a_j^{(l,s)}| - \mu_{\Delta}}{\sigma_{\Delta}} > \zeta \right\}
\label{eq:safety_neurons}
\end{equation}
where $\mu_{\Delta}$ and $\sigma_{\Delta}$ are the mean and standard deviation of $\{|\Delta a_j^{(l,s)}|\}_{j=1}^{d}$.

Safety-critical neurons constitute a sparse subset (${\sim}3\%$) of total neurons~\cite{chen2025towards, wu2026gatebreaker}, creating a concrete vulnerability exploitable by pruning a small fraction of neurons.

\section{Representational Analysis}
\label{app:interp_supplement}

As a supplement to \autoref{subsec:activation_effects}, we verify that \OurModel preserves representational structure through two complementary analyses across all four architectures.

\begin{figure*}[t]
\centering
\includegraphics[width=\textwidth]{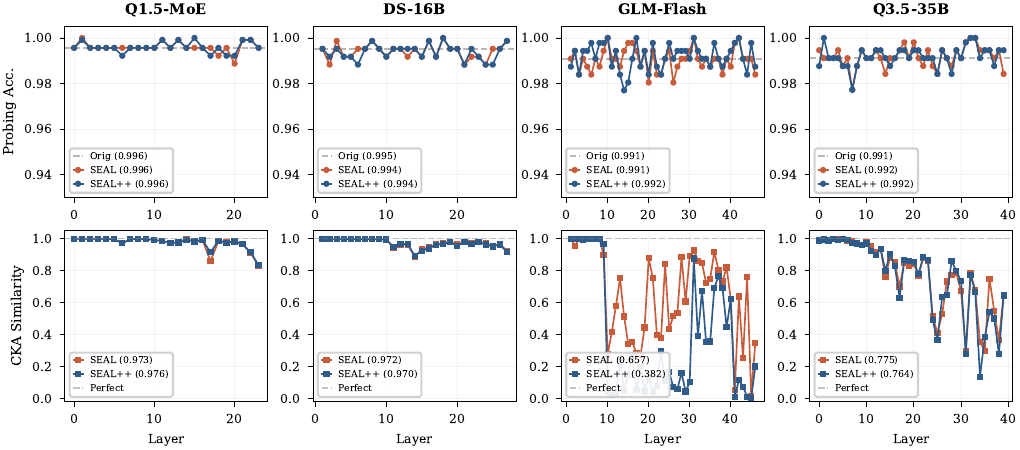}
\caption{Per-layer interpretability analysis across four architectures. Top: linear probing accuracy for harmful-versus-benign classification. Bottom: CKA similarity between base and aligned representations. All methods preserve safety concept separability.}
\Description{Eight-panel grid showing per-layer linear probing accuracy (top) and CKA similarity (bottom) for four MoE architectures under three training conditions.}
\label{fig:interp_probing_cka}
\end{figure*}

\paragraph{Safety concept separability.}
\label{app:linear_probing}
Per-layer linear probes (logistic regression on hidden states) achieve $>$96\% harmful-versus-benign classification accuracy across all layers and all conditions (\autoref{fig:interp_probing_cka}, top row). Average probe accuracy changes by less than 0.2\% after \OurModel training on every architecture, confirming that safety concepts remain linearly decodable from shared expert representations.

\paragraph{Representational similarity.}
\label{app:cka}
CKA~\cite{pmlr-kornblith19a-cka} between the base and aligned models (\autoref{fig:interp_probing_cka}, bottom row) remains high on Qwen1.5 (0.97) and DeepSeek (0.97). GLM (0.38--0.66) and Qwen3.5 (0.76--0.78) show lower CKA, reflecting more extensive representational restructuring on deeper architectures with concentrated safety encoding. Despite this drift, probe accuracy remains near-perfect, indicating that the
defence reorganises the representational structure on these
architectures without erasing the safety separation it depends on.

\begin{figure*}[!t]
\centering
\includegraphics[width=\textwidth]{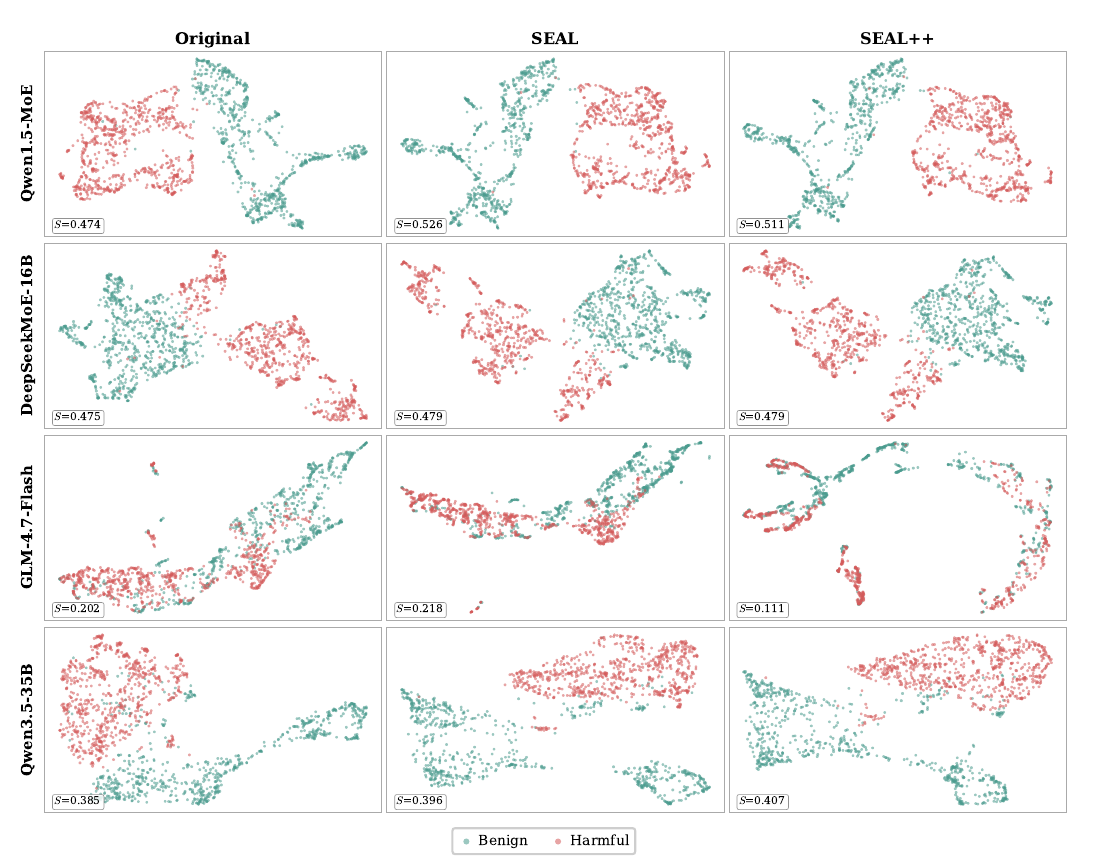}
\caption{UMAP projections of shared expert activations across four architectures and three conditions, $S$ = silhouette score.}
\Description{Four-by-three grid of UMAP scatter plots showing harmful and benign
activation clusters for four MoE architectures under original, SEAL, and SEAL++
conditions.}
\label{fig:umap_grid}
\end{figure*}

\paragraph{Embedding visualization.}
\label{app:umap_embedding}
UMAP projections (\autoref{fig:umap_grid}) show that harmful-benign cluster structure is maintained or improved after training across all four architectures (silhouette scores stable or increasing), confirming that defense training does not collapse the activation space.

\begin{figure*}[!t]
\centering
\includegraphics[width=0.8\textwidth]{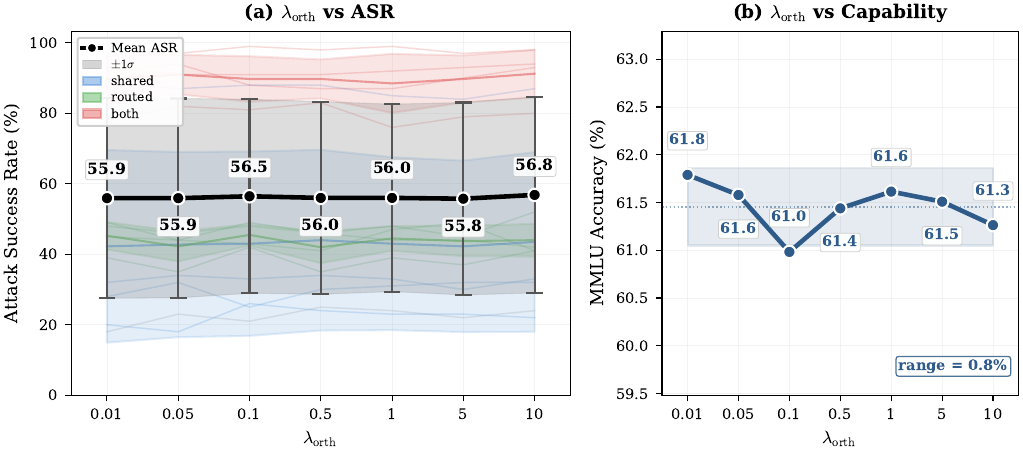}
\caption{Sensitivity of pruning attack ASR and MMLU accuracy to $\lambda_{\text{orth}}$ on Qwen1.5-MoE (500-step proxy). Semi-transparent lines show individual attack configurations colored by scope. Bold black line with error bars indicates mean $\pm 1\sigma$. Both metrics remain constant across three orders of magnitude, confirming low sensitivity.}
\Description{Two-panel figure showing ASR and MMLU accuracy across seven lambda values.}
\label{fig:ortho_sweep}
\includegraphics[width=0.95\textwidth]{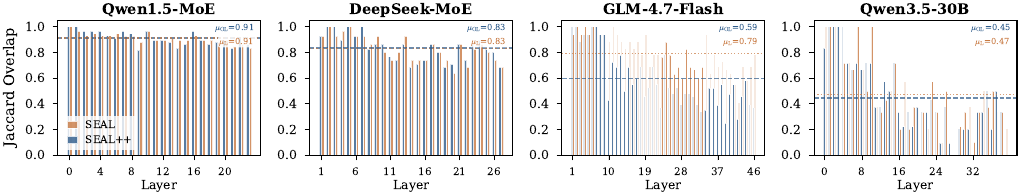}
\caption{Per-layer Jaccard overlap of safety neuron identity (pre- vs.\ post-training) across four architectures. Dashed lines indicate model-wide means. Models with concentrated shared-expert safety encoding (Qwen1.5, DeepSeek) preserve $>$80\% of neuron identity; Qwen3.5 shows substantially more redistribution.}
\Description{Bar charts showing per-layer Jaccard overlap of safety neuron sets before and after training for four architectures.}
\label{fig:dispersion}
\end{figure*}

\paragraph{Safety-neuron identity preservation.}
\label{app:dispersion}
Beyond the UMAP cluster structure, we measure per-layer Jaccard overlap between pre-training and post-training safety neuron sets across all four architectures
(\autoref{fig:dispersion}). Qwen1.5 maintains the highest overlap (mean $J = 0.91$), followed by DeepSeek ($J = 0.83$) and GLM ($J = 0.59$--0.79, with \OurMethodA preserving more neuron identity than \OurMethodB). Qwen3.5 shows the lowest overlap ($J = 0.45$--0.47).

\section{Computational Overhead}
\label{app:overhead}

\autoref{tab:overhead} reports the computational and storage costs of \OurModel. The orthogonal constraint (\autoref{eq:orth_loss}) adds roughly $35$\,GFLOPs per training step for Qwen1.5-MoE and $20$\,GFLOPs for DeepSeekMoE, under $0.02\%$ of the DPO forward-backward pass. Trainable parameters constitute $0.22$--$0.25\%$ of total model parameters. At inference time the merged adapter
introduces zero additional FLOPs because LoRA weights are folded into the base parameters. The safety-subspace bases $\mat{V}_{\text{safe}}^{(l,s)} \in \mathbb{R}^{d \times r_s}$ require at $21.2$\,MB of storage at most, enabling lightweight deployment alongside the adapter checkpoint.

\begin{table}[!t]
\centering
\caption{Computational overhead of \OurModel on Qwen1.5 and DeepSeek. GLM and Qwen3.5 overhead is proportional to their shared expert module count (\autoref{tab:param_cost}).}
\label{tab:overhead}
\footnotesize
\begin{tabular}{@{}l r r@{}}
\toprule
& \textbf{Qwen1.5-MoE-A2.7B} & \textbf{DeepSeekMoE-16B} \\
\midrule
Total / active params & 14.3B / 2.7B & 16.4B / 2.8B \\
Trainable params & 35.4M (0.25\%) & 35.8M (0.22\%) \\
LoRA rank $r$ / safety dim.\ $r_s$ & 64 / 32 & 64 / 32 \\
Modules $|\mathcal{M}|$ & 72 & 162 \\
\midrule
Orth.\ constraint FLOPs/step & 35\,GFLOPs & 20\,GFLOPs \\
vs.\ DPO fwd+bwd             & $<$0.02\%  & $<$0.02\%  \\
\midrule
Adapter storage (BF16)       & 70.8\,MB   & 71.6\,MB   \\
Safety subspace storage      & 9.4\,MB    & 21.2\,MB   \\
\bottomrule
\end{tabular}
\end{table}

\section{Per-Benchmark Capability Results}
\label{app:capability_full}

\autoref{tab:capability_full} reports individual benchmark scores for all models and methods.

\begin{table}[!htb]
\centering
\caption{Per-benchmark capability (\%$\uparrow$) for all models and defense methods. TQA\,$=$\,TruthfulQA, PQA\,$=$\,PopQA. Average capability varies by at most 1.8\% from Original.}
\label{tab:capability_full}
\setlength{\tabcolsep}{3pt}
\small
\begin{tabular}{@{}ll rrrrr r@{}}
\toprule
\textbf{Model} & \textbf{Method} & \textbf{GSM8K} & \textbf{MMLU} & \textbf{ARC} & \textbf{TQA} & \textbf{PQA} & \textbf{Avg} \\
\midrule
\multirow{3}{*}{\textit{Qwen1.5}} & Original & 50.3 & 63.2 & 77.5 & 33.2 & 24.7 & 49.8 \\
& \OurMethodA & 49.4 & 62.3 & 77.4 & 35.0 & 21.9 & 49.2 \\
& \OurMethodB & 48.2 & 62.0 & 76.9 & 35.6 & 21.3 & 48.8 \\
\midrule
\multirow{3}{*}{\textit{DeepSeek}} & Original & 45.5 & 50.2 & 55.4 & 33.3 & 25.0 & 41.9 \\
& \OurMethodA & 44.4 & 50.7 & 54.8 & 40.4 & 25.1 & 43.1 \\
& \OurMethodB & 45.7 & 49.6 & 54.4 & 40.1 & 24.7 & 42.9 \\
\midrule
\multirow{3}{*}{\textit{GLM}} & Original & 77.8 & 70.7 & 81.0 & 35.7 & 27.1 & 58.5 \\
& \OurMethodA & 80.4 & 70.9 & 79.7 & 41.9 & 28.5 & 60.3 \\
& \OurMethodB & 81.7 & 69.9 & 79.5 & 41.1 & 23.1 & 59.1 \\
\midrule
\multirow{3}{*}{\textit{Qwen3.5}} & Original & 95.1 & 84.3 & 94.7 & 36.5 & 32.5 & 68.6 \\
& \OurMethodA & 93.6 & 84.1 & 94.8 & 44.9 & 29.7 & 69.4 \\
& \OurMethodB & 92.6 & 83.5 & 94.5 & 45.5 & 30.3 & 69.3 \\
\bottomrule
\end{tabular}
\end{table}

\section{Sensitivity Analysis}
\label{app:sensitivity}

We study the orthogonal-constraint weight $\lambda_{\text{orth}}$ from two
complementary angles: a full-training ablation at two anchor values
($0.1$ vs $1.0$) (\autoref{tab:ablation_lambda}), and a finer-grained sweep over seven values spanning three orders of magnitude ($0.01$--$10.0$) on a 500-step Qwen1.5-MoE training proxy (\autoref{fig:ortho_sweep}).

\paragraph{Full-training ablation.}
Raising $\lambda_{\text{orth}}$ from 0.1 to 1.0 produces architecture-dependent
effects. On DeepSeek, the stronger constraint substantially reduces
shared-scope ASR (from 21.5\% to 8.8\%) with marginal changes on other scopes,
suggesting that the orthogonal penalty helps concentrate safety
representations in the shared-expert subspace. On Qwen1.5, changes across all
scopes remain within 4\% in either direction. Neither model shows meaningful
capability impact.

\begin{table}[!t]
\centering
\caption{Orthogonal constraint weight ablation ($\lambda_{\text{orth}}$): worst-case ASR (\%$\downarrow$) and capability average (\%$\uparrow$). Reported for Qwen1.5 and DeepSeek.}
\label{tab:ablation_lambda}
\setlength{\tabcolsep}{3pt}
\footnotesize
\begin{tabular}{@{}lc cccc c@{}}
\toprule
& & \multicolumn{4}{c}{\textbf{Pruning ASR (\%$\downarrow$)}}
& \textbf{Capability} \\
\cmidrule(lr){3-6} \cmidrule(lr){7-7}
\textbf{Model} & $\lambda$
  & \textbf{Base}
  & \textbf{Shared}
  & \textbf{Routed}
  & \textbf{Both}
  & \textbf{Avg} \\
\midrule
& 0.1 & 13.1 & 18.3 & 13.9 & 16.6 & 48.8 \\
\multirow{-2}{*}{\textit{Qwen1.5-MoE-A2.7B}} & 1.0 & 14.3 & 22.3 & 13.3 & 19.7 & 49.1 \\
\midrule
& 0.1 & 9.9 & 21.5 & 14.0 & 17.3 & 42.9 \\
\multirow{-2}{*}{\textit{DeepSeekMoE-16B}} & 1.0 & 9.3 & 8.8 & 16.2 & 16.0 & 43.3 \\
\bottomrule
\end{tabular}
\end{table}

\paragraph{Fine-grained sweep.}
\autoref{fig:ortho_sweep} extends the ablation across seven values on
Qwen1.5-MoE using a 500-step training proxy (26\% of the full training run).
The left panel plots per-configuration pruning-attack ASR across all three
scopes, with the bold black line indicating the cross-configuration mean. The
right panel tracks MMLU accuracy. Both metrics remain
effectively constant across the full range, confirming that the defense is
insensitive to the constraint weight within the evaluated interval and
supporting our default choice of $\lambda_{\text{orth}} = 0.1$.

\section{Training Dynamics of Routed and Shared Experts}
\label{app:training_dynamics}

\begin{center}
\includegraphics[width=\columnwidth]{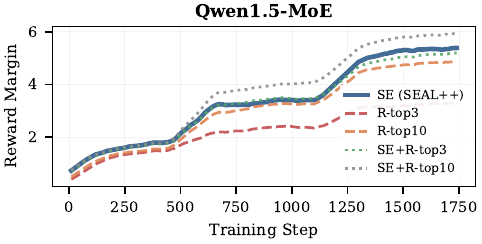}
{\captionsetup{hypcap=false}%
\captionof{figure}{DPO reward margin during training on Qwen1.5. Conditions that include shared experts (SE, SE+R-top3, SE+R-top10) achieve high margins ($>$5.0), while routed-only conditions (R-top3, R-top10) plateau much lower, consistent with their weaker defense in \autoref{tab:rasa_lite}.}
\label{fig:training_dynamics}
\Description{Single-panel line chart showing reward margin over 1932 training steps for five conditions on Qwen1.5.}
}
\end{center}

\autoref{fig:training_dynamics} compares reward margin (the DPO preference gap between chosen and rejected responses) across shared-expert (SE) and routed-expert (RE) training conditions on Qwen1.5. SE maintains consistently higher margins throughout training, reaching a final value of 5.3 while R-top3 plateaus at 3.3. R-top3, which trains only routed experts with a matched parameter budget, develops margins 38\% lower than SE, explaining its substantially weaker defense in \autoref{tab:rasa_lite}. Joint SE+R conditions track SE closely, confirming that the shared expert component drives margin growth.

\end{document}